\documentclass[]{interact}

\usepackage{epstopdf}
\usepackage[caption=false]{subfig}

\usepackage[natbibapa,nodoi]{apacite}
\theoremstyle{plain}

\theoremstyle{definition}

\theoremstyle{remark}

\usepackage{multirow}
\usepackage{verbatim}
\usepackage{makecell}
\usepackage{longtable}
\usepackage{pdflscape}
\usepackage{array}
\usepackage{xcolor}

\begin{document}

\articletype{RESEARCH ARTICLE}

\title{PaReGTA: A Temporally Aware LLM-Based Patient Representation Framework for EHR Analytics}

\author{
\name{Kihyuk Yoon\textsuperscript{a, b},
Lingchao Mao\textsuperscript{a},
Catherine Chong\textsuperscript{c}, 
Todd Schwedt\textsuperscript{c},
Chia-Chun Chiang\textsuperscript{c}, 
Jing Li\textsuperscript{a}*}
\thanks{*Corresponding author. Email: jli3175@gatech.edu}
\affil{\textsuperscript{a}Georgia Institute of Technology, \textsuperscript{b}UNIST, \textsuperscript{c}Mayo Clinic}
}

\maketitle

\begin{abstract}
Temporal information in structured electronic health records (EHRs) is often lost in sparse one-hot or count-based representations, while sequence models can be costly and data-hungry. We propose PaReGTA, an LLM-based encoding framework that (i) converts longitudinal EHR events into visit-level templated text with explicit temporal cues, (ii) learns domain-adapted visit embeddings via lightweight contrastive fine-tuning of a sentence-embedding model, and (iii) aggregates visit embeddings into a fixed-dimensional patient representation using hybrid temporal pooling that captures both recency and globally informative visits. The resulting fixed-dimensional patient representations can be used with conventional downstream machine-learning models. To examine factor-level sensitivity, we use PaReGTA-RSS (Representation Shift Score), a prespecified factor-removal analysis that recomputes patient representations after removing clinically defined factor groups and quantifies the resulting change in the fitted logit of a fixed logistic-regression model. We evaluated PaReGTA in a cohort of 39,088 patients with migraine from the All of Us Research Program (AoU) on a retrospective patient-level classification task with an EHR-derived target. In an exploratory comparison on the fixed held-out test cohort of 7,818 patients, PaReGTA-Gap + LightGBM, used as a post hoc analytical reference, had the highest reported AUC, accuracy, and F1 values among the evaluated sparse, BERT-based EHR, and recurrent approaches in this cohort.
\end{abstract}

\begin{keywords}
Electronic Health Records; Large Language Models; Temporal Information; Patient Representation; Migraine
\end{keywords}

\section{Introduction}

Electronic health records (EHRs) capture diagnoses, prescriptions, test results, comorbidities, and patient-reported outcomes over extended periods, often spanning years \citep{shickel_deep_ehr_2018,Wang2024RecentAdvances}. This longitudinal structure provides a unique opportunity to model disease trajectories, forecast clinical progression and healthcare utilization, evaluate real-world treatment effects, and analyze care processes across diverse populations \citep{rajkomar_scalable_2018,miotto_deeppatient_2016}. However, realizing these benefits depends critically on how to encode the EHR data, particularly on whether the encoding strategy preserves temporal information in a form that downstream models can effectively leverage \citep{shickel_deep_ehr_2018}.

Despite the inherent sequential nature of EHRs, many practical prediction pipelines still rely on simplified patient-level representations that discard the temporal information in EHRs. Common approaches include one-hot encodings and aggregated count vectors constructed over fixed observation windows \citep{shickel_deep_ehr_2018,steinberg_clmbr_2020}. These representations are attractive because they are standardized and integrate easily with conventional machine learning models. However, they collapse visit-level records into unordered summaries and discard clinically meaningful temporal dynamics. As a result, distinct clinical narratives (e.g., the emergence of one disease followed by another) may become indistinguishable, even when their prognostic implications differ. This limitation is repeatedly noted as a central barrier to fully leveraging EHRs for longitudinal modeling and decision support \citep{shickel_deep_ehr_2018}.

Prior work has primarily explored two directions for preserving temporal information. The first direction is time-aware feature engineering, where temporal dynamics are approximated through hand-crafted variables such as counts within predefined windows, recency indicators, time-since-last-event measures, or gap-based features \citep{shickel_deep_ehr_2018,tang_fiddle_2020}. Although widely adopted, this approach has several inherent limitations. It requires substantial manual effort and domain expertise to design clinically meaningful temporal features, making it difficult to scale across different clinical contexts \citep{bengio2013representation,miotto_deeppatient_2016}. The hand-crafted features are inherently task-specific and may not generalize across diseases or prediction objectives \citep{lasko2013computational,choi_med2vec_2016}. Moreover, predefined aggregation windows (e.g., 30-day, 90-day counts) impose rigid temporal granularity, potentially losing fine-grained patterns that fall outside these boundaries or masking clinically relevant variations within them \citep{lipton2016directly,xiao_dl_ehr_review_2018}. Feature selection is also inherently subjective, which may reduce reproducibility and comparability across studies \citep{johnson2016machine}. Simple aggregation-based features struggle to capture complex temporal dependencies such as nonlinear trends, variable-length patterns, or multi-scale interactions between events  \citep{che_grud_2018,baytas_tlstm_2017}.

The second direction is sequence modeling, which directly encodes visit sequences using Recurrent Neural Networks (RNNs), attention mechanisms, or related architectures. Representative approaches include recurrent models for predicting when the next visit will occur \citep{choi_doctorai_2016}, memory-augmented approaches for irregular long-term dependencies \citep{pham_deepcare_2016}, convolutional architectures that learn motifs from episodic records \citep{nguyen_deepr_2017}, attention-based models for diagnosis prediction \citep{ma_dipole_2017}, and interpretable attention mechanisms for visit- and code-level attributions \citep{choi_retain_2016}. Handling irregular sampling and missingness often requires additional architectural machinery \citep{baytas_tlstm_2017,che_grud_2018}. Although sequence-based models can represent temporal information more faithfully, they often incur higher computational and deployment costs and can be sensitive to data sparsity and the heterogeneity common in real-world EHRs. 

Transformer-based approaches further enhance modeling capacity for long-range dependencies \citep{vaswani_attention_2017} and have been adapted to structured EHR sequences \citep{li_behrt_2020,rasmy_medbert_2021}. In particular, CEHR-BERT introduced artificial time tokens inserted between visits to explicitly encode temporal intervals within the BERT framework \citep{Pang2021CEHRBERT}, demonstrating that temporally aware tokenization can improve prediction performance over standard BERT adaptations \citep{Pang2021CEHRBERT}. In parallel, autoregressive EHR foundation models have been proposed to improve robustness and mitigate temporal distribution shift \citep{steinberg_clmbr_2020,guo_ehr_foundation_shift_2023}. A recent survey of these deep learning approaches \citep{Wang2024RecentAdvances} highlights persistent challenges, including temporal irregularity and data heterogeneity. These advances highlight the promise of representation learning for temporally structured EHRs, but practical barriers to clinical adoption remain non-trivial: long sequences, irregular timing, site-specific vocabularies, and the need to maintain specialized deep architectures beyond conventional clinical machine learning pipelines \citep{shickel_deep_ehr_2018}.

Beyond model design considerations, data normalization of semi-structured EHR tables is a practical barrier. In many EHR systems, medication records are stored at the product name level with heterogeneous naming conventions. Standardization through vocabularies such as RxNorm and common data models such as OMOP/OHDSI \citep{nelson_rxnorm_2011,hripcsak_ohdsi_2015}  improves interoperability, yet consistent enforcement at the point of data entry is difficult in practice. Each institution typically maintains only limited-scale datasets and cross-institution inconsistencies in coding practices can hinder harmonization. Recent work on embedding harmonization has attempted to address this issue by aligning medical event representations learned from different data sources into a common embedding space, though such approaches still require shared vocabularies or correspondences between sources \citep{Li2024EmbeddingHarmonization}. Retrospective standardization of heterogeneous categorical laboratory results can require dedicated mapping procedures \citep{park2019salt_standardization}. Transitions between EHR systems can also complicate the preservation and retrieval of legacy records \citep{reider2020ehr_transitions}. Consequently, obtaining large-scale EHR datasets with uniformly standardized concepts remains challenging. These data constraints further limit the use of deep learning architectures, which typically require substantial volumes of consistently normalized data to achieve robust performance. When data are fragmented across sites, inconsistently mapped, or restricted to small cohorts, these data requirements become a practical barrier to deploying deep models in real-world clinical settings. 

Large language models (LLMs) and sentence-embedding methods offer an alternative encoding route. Because LLMs are pretrained on massive and diverse text corpora, they encode broad semantic knowledge that can generalize across heterogeneous terminologies and naming conventions. This built-in knowledge allows heterogeneous clinical records to be converted into embedded dense vectors, reducing dependence on rigid code taxonomies and manual feature design \citep{lee_biobert_2020,alsentzer_clinicalbert_2019,yang_gatortron_2022,reimers_sbert_2019,gao_simcse_2021}. Early work demonstrated that converting medical codes into text descriptions enables unified EHR representations across heterogeneous coding systems \citep{Hur2021TextBasedCode}. More recently, general-purpose LLM embedding models have been evaluated as encoders of longitudinal EHR data across multiple clinical prediction tasks \citep{Hegselmann2026LLMEHREncoders}, while GRASP combines LLM-derived medical-concept embeddings with a Transformer to support the transfer of EHR-based predictions across countries and coding systems \citep{Kirchler2026GRASP}. Recent studies also explore prompting strategies and transforming tabular EHR into ``pseudo-notes'' \citep{zhu2024promptinglargelanguagemodels} and discuss opportunities and caveats for language-model-based risk prediction \citep{acharya_clinical_risk_lm_2024}. 

However, several methodological challenges remain for modeling structured longitudinal EHR: (i) how to textualize EHR data in a clinically meaningful way while preserving temporal information, (ii) how to adapt embedding models to cohort-specific vocabularies without requiring costly labeled sentence pairs, and (iii) how to aggregate multiple visit-level embeddings into a patient-level representation that preserves temporal cues while remaining compatible with conventional machine-learning pipelines \citep{shickel_deep_ehr_2018,reimers_sbert_2019,gao_simcse_2021}. 

Furthermore, prediction performance alone is insufficient for clinical adoption: models must support meaningful explanations. Yet LLM-based encoding makes conventional feature-importance analysis difficult. Model-intrinsic measures (e.g., linear coefficients) correspond to latent embedding dimensions after multiple stages of text construction, sentence encoding, and pooling, and therefore cannot be interpreted directly in terms of clinical variables. Perturbation-based tools such as SHAP \citep{lundberg_shap_2017} or LIME \citep{ribeiro_lime_2016} are also challenging here: applying them faithfully would require repeatedly perturbing the raw clinical factor, re-running the entire text→embedding→pooling pipeline, and recomputing predictions, which is expensive and sensitive to perturbation design \citep{lundberg_shap_2017,ribeiro_lime_2016}. 

In this study, we propose PaReGTA (Patient Representation Generation with Temporal Aggregation), a modular LLM-based representation framework for longitudinal EHR datasets. PaReGTA converts structured EHR events into templated visit-level sentences, fine-tunes a sentence-embedding model using a contrastive objective \citep{reimers_sbert_2019,gao_simcse_2021}, and aggregates visit embeddings into a patient-level representation using a temporally aware pooling scheme that emphasizes both recency and globally important visits. This hierarchical design is grounded in established principles for learning from sets and hierarchically structured sequences through pooling/attention mechanisms \citep{zaheer_deepsets_2017,yang_han_2016}. To support factor-level inspection of the fitted pipeline, we further introduce PaReGTA-RSS (Representation Shift Score), a prespecified factor-removal sensitivity analysis that quantifies changes in the fitted logit of a fixed model after clinically defined factor groups are removed, allowing patient-level changes and cohort-level summaries to be examined.  

Recent LLM-based EHR representation approaches differ in both their representation unit and their treatment of longitudinal information. In its default representation, \citet{Hegselmann2026LLMEHREncoders} serializes each patient record as a list of natural-language code descriptions and retains only the most recent occurrence of each code. GRASP precomputes embeddings of the natural-language names of condition, procedure, and drug concepts, retains repeated concept occurrences as embedding tokens, and applies a Transformer to the resulting variable-length collection \citep{Kirchler2026GRASP}. Consequently, GRASP represents frequency implicitly through token multiplicity, but its current implementation omits event timing and sequence order and does not explicitly preserve visit-level temporal structure. PaReGTA instead constructs text at the visit level, incorporates temporal cues through the ``Date,'' ``Gap,'' or ``Month'' prefixes, encodes each visit using a sentence-embedding model, and aggregates the resulting visit-level embeddings through temporal pooling. Thus, PaReGTA differs from these approaches primarily in its representation unit and explicit treatment of visit-level temporal structure rather than in introducing a new language-model architecture.

We validate PaReGTA on the All of Us (AoU) Research Program \citep{denny_allofus_2019}, one of the largest harmonized, multi-institutional EHR datasets available for biomedical research. AoU aggregates longitudinal and structured EHR data from hospitals, clinics, and home-based clinical care settings across the United States under a harmonized common data model. Leveraging this large-scale dataset, we evaluate PaReGTA as a framework for generating patient-level representations on a retrospective patient-level classification task with an EHR-derived migraine target, demonstrating its effectiveness as a temporally aware patient-representation layer for longitudinal EHR analytics under real-world heterogeneity in clinical records. 

In summary, this work makes the following contributions:
\begin{itemize}
    \item \textbf{EHR preprocessing that incorporates temporal information while reducing reliance on prior knowledge.} After disease-specific cohort construction, PaReGTA orders the supplied patient-visit rows chronologically and converts the intervals between retained visits into text according to fixed rules. This approach minimizes the need for extensive prior knowledge or repetitive manual feature engineering while preserving both the recorded clinical concepts and temporal information in the constructed longitudinal EHR table.
    \item \textbf{A reproducible patient-level EHR analysis framework and controlled comparative evaluation.} We encoded the generated visit-level text with a sentence-embedding model, applied temporal pooling to aggregate the visit-level embeddings into a single patient-level representation, and linked this representation to downstream machine learning. We also compared PaReGTA with one-hot encoding, BoC, MedBERT, CEHR-BERT, T-LSTM, and RETAIN on the same fixed held-out test cohort. This evaluation shows how temporal information introduced through deterministic textualization can be incorporated into patient-level representations through sentence encoding and temporal pooling.
    \item \textbf{Patient-level findings in migraine using PaReGTA.} Using PaReGTA, we examined not only predictive performance but also patient-level medication, comorbidity, and temporal patterns associated with the EHR-derived target. Performance remained close to that of the full-input model when medication records were excluded and only comorbidity records were used, indicating that accumulated comorbidity records also contained information relevant to distinguishing target-positive and target-negative patients under the evaluated conditions. In the factor-removal analysis, the magnitude of fitted-logit changes differed across predefined factor groups and sex-at-birth subsets. Taken together, these analyses provide a retrospective characterization of how the fitted pipeline responded to the evaluated EHR inputs in this cohort.
\end{itemize}

\begin{table}[t]
\tbl{Constructed example of the AoU EHR data structure. Because AoU does not permit the export of participant-level records, all values shown are synthetic and are provided only to illustrate preprocessing and textualization.}
{\begin{tabular}{ccccccc} 
\toprule
ID & Age & Sex & Race & Visit date & Medication & Diagnosis \\
\midrule
1 & 33 & Male & Asian & 2021-06-01 & - & Unspecified migraine \\
1 & 33 & Male & Asian & 2021-06-01 & \makecell{acetaminophen 325 MG \\ Oral Tablet} & - \\
1 & 33 & Male & Asian & 2021-06-03 & - & Depression \\
\midrule
1 & 33 & Male & Asian & 2021-07-01 & - & Flu \\
1 & 33 & Male & Asian & 2021-07-02 & \makecell{acetaminophen 325 MG \\ Oral Tablet} & - \\
\midrule
1 & 33 & Male & Asian & 2021-09-01 & - & Migraine without Aura \\
1 & 33 & Male & Asian & 2021-09-01 & - & Depression \\
1 & 33 & Male & Asian & 2021-09-02 & \makecell{lasmiditan 100 MG \\ Oral Tablet} & - \\
1 & 33 & Male & Asian & 2021-09-02 & - & Insomnia \\
\midrule
1 & 33 & Male & Asian & 2021-12-01 & - & \makecell{Chronic Migraine \\ without Aura} \\
1 & 33 & Male & Asian & 2021-12-01 & \makecell{lasmiditan 100 MG \\ Oral Tablet} & - \\
1 & 33 & Male & Asian & 2021-12-02 & ibuprofen & - \\
1 & 33 & Male & Asian & 2021-12-01 & - & Depression \\
1 & 33 & Male & Asian & 2021-12-01 & - & Insomnia \\
\bottomrule
\end{tabular}}
\label{table_example_data}
\end{table}
\section{Methods}\label{s:4:methods}

\subsection{Input data requirements and illustrative preprocessing example}

PaReGTA receives a disease-specific longitudinal table indexed by patient and visit date, with medication and comorbidity content stored in separate row-level fields. It does not define the visit dates, construct the disease cohort, link raw EHR events to visit records, or define the downstream target. The following synthetic migraine example illustrates the patient-visit input format expected by PaReGTA. The migraine-specific procedures used to construct this case-study dataset are not components of the general PaReGTA framework and are documented separately in Appendix~\ref{app:migraine_reference_dates}.

\begin{table}[t]
\tbl{Simplified example of the preprocessed AoU cohort table.}
{\begin{tabular}{cccccccc} 
\toprule
ID & Age & Sex & Race & Visit date & Medication & Comorbidities & Migraine type \\
\midrule
1 & 33 & Male & Asian & 2021-06-01 & \makecell{acetaminophen \\ 325 MG Oral Tablet} & Depression & Episodic \\
\midrule
1 & 33 & Male & Asian & 2021-09-01 & \makecell{lasmiditan 100 MG \\ Oral Tablet} & \makecell{Depression, \\ Insomnia} & Episodic \\
\midrule
1 & 33 & Male & Asian & 2021-12-01 & \makecell{lasmiditan 100 MG \\ Oral Tablet, \\ ibuprofen} & \makecell{Depression, \\ Insomnia} & Chronic \\
\bottomrule
\end{tabular}}
\label{table_example_data2}
\end{table}

Table~\ref{table_example_data} provides a simplified illustration of the synthetic source longitudinal EHR records used in the migraine case study. Table~\ref{table_example_data2} illustrates the visit-level records produced by the case-study data-preparation procedure before PaReGTA textualization. The values in the Migraine type column represent the migraine type associated with each visit record during case-study preprocessing and do not define the patient-level target. We defined the patient-level target independently using the ever-chronic rule described in the following subsection.

\subsubsection{Patient-level target definition}
\label{subsec:patient_level_target}

We defined a binary patient-level target from each patient's available condition history. The target was set to 1 (target-positive) when \texttt{Chronic\_Migraine\_without\_Aura} occurred at least once, including in a record without a usable date, and to 0 (target-negative) otherwise. We selected this operational ever-chronic rule to align target assignment as closely as possible with the longitudinal migraine coding available in the EHR. When qualifying chronic and non-chronic migraine records co-occurred, the presence of a chronic record determined the patient-level target.

For the longitudinal code-history audit, we used available record dates to assign each patient to one of five mutually exclusive descriptive categories: non-chronic-only, chronic-only, one-way non-chronic-to-chronic code history, one-way chronic-to-non-chronic code history, or complex mixed history. We assigned a mixed history to the complex category when chronic and non-chronic records occurred on the same date, the recorded category changed direction more than once, or missing dates prevented determination of the order. Dates were required to characterize record order but not for patient-level target assignment. 

For cohort-level descriptive analyses, we defined each patient's visit count as the number of unique normalized dates in the retained analytic table. We defined the observation span as the interval between each patient's earliest and latest retained dates and divided the interval in days by 30.4375 to express it in months. Patients with only one retained date therefore had an observation span of 0 months. We summarized visit counts and observation spans for the full analytic cohort and separately for the mutually exclusive target-negative and target-positive groups using medians, means, standard deviations (SDs), and ranges.

We identified migraine and comorbidity records using prespecified operational ICD-9-CM and ICD-10-CM concept sets. The complete migraine and comorbidity code lists are reported in Appendix Tables~\ref{tab:migraine_icd_codes} and~\ref{tab:comorbidity_icd_codes}, respectively.

\subsection{PaReGTA framework}
\label{subsection_representation_generation}

\subsubsection{Framework overview}

\begin{figure}
\centering
\includegraphics[width=0.99\columnwidth]{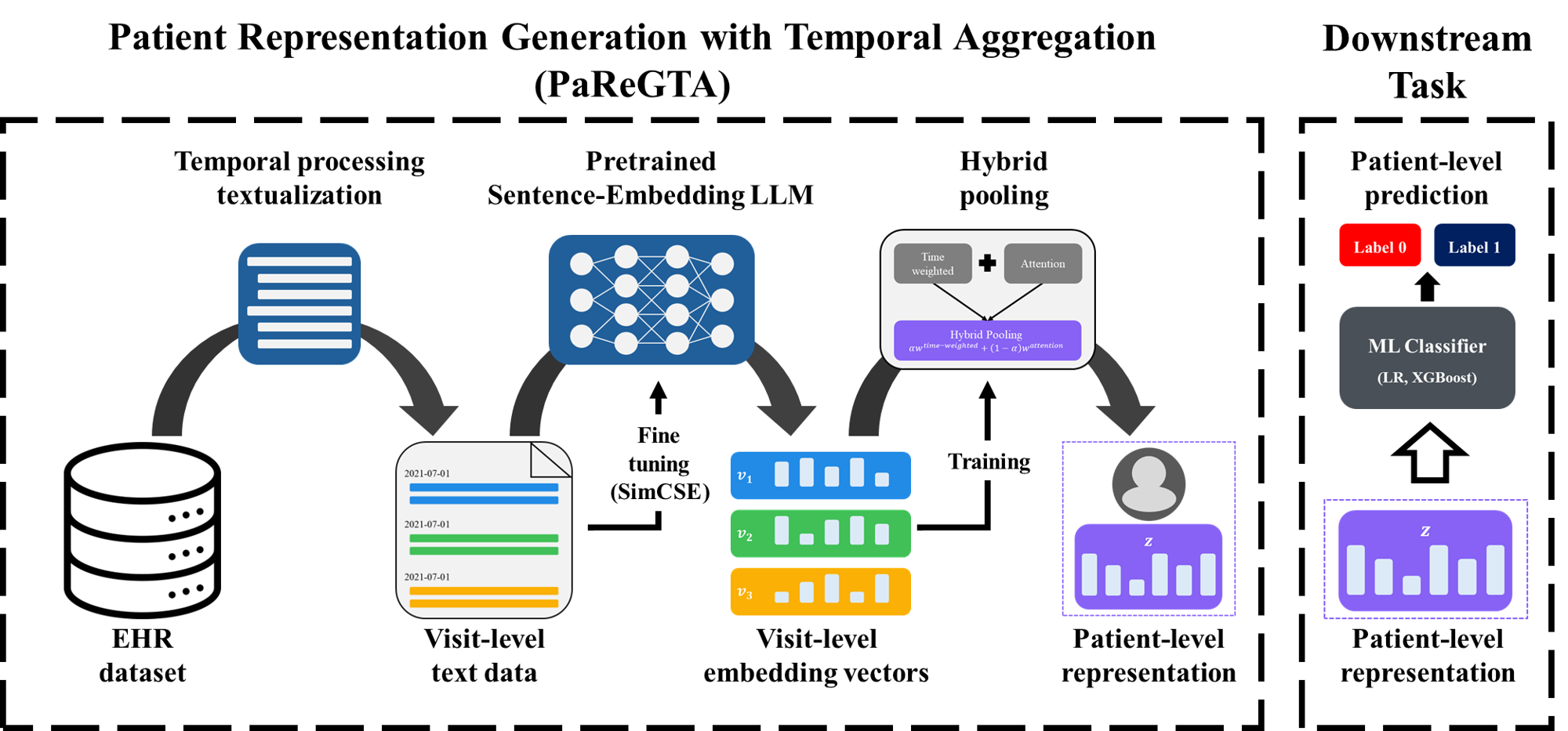}
\caption{Overview of PaReGTA.}
\label{fig_llm_pipeline}
\end{figure}

Given a constructed cohort as input, the goal of our approach, PaReGTA, is to transform it into fixed-dimensional patient representations while preserving the natural patient→visit→event hierarchy and temporal cues and remaining compatible with standard downstream machine-learning models. To achieve the goal, PaReGTA comprises three main stages: (i) \textbf{Visit-level text construction}: a deterministic textualization procedure orders the supplied patient-visit rows and applies fixed templates to their retained medication and comorbidity fields; (ii) \textbf{Domain adaptation via contrastive learning}: we obtain visit-level embeddings using a pretrained sentence-embedding LLM and perform lightweight domain adaptation via SimCSE fine-tuning on cohort-specific visit texts; and (iii) \textbf{Hybrid temporal pooling into patient representations}: visit-level embeddings of each patient are aggregated into a single patient-level representation using hybrid temporal pooling that combines time-decay weighting to emphasize recency and attention-based weighting to emphasize globally informative visits. The resulting patient representations are used as inputs to conventional classifiers for downstream analytical and classification tasks in this study. An overview is provided in Figure \ref{fig_llm_pipeline}.

\subsubsection{Visit-level textualization of EHR dataset}

During disease-specific preprocessing, medication and comorbidity events associated with the same constructed visit were organized within the corresponding patient-visit row. PaReGTA textualization began from this preprocessed table, sorted the supplied rows by patient and visit date to calculate temporal prefixes, and converted the medication and comorbidity fields already attached to each row into separate domain-specific strings. It therefore did not perform a second same-date merge or reassign raw EHR events to visit records. Within each domain, we preserved the stored record order and did not apply alphabetical or clinically motivated reordering.

For medication events, we constructed product-level strings from retained standardized concept names, associated amount fields, and medication-class labels, and joined multiple entries with a fixed semicolon separator. If one domain was empty while the other contained a retained event, we preserved the visit and represented the empty domain with the literal lowercase token \texttt{none} rather than an empty string; visits with neither medication nor comorbidity content were excluded upstream and therefore did not enter textualization or sentence encoding. The token \texttt{none} indicates only that no record was retained for that domain at the corresponding visit and should not be interpreted as confirmed clinical absence or the absence of a prescription. We did not infer or impute medications that were not documented in the EHR-derived input.

We then applied fixed templates that differed in their temporal prefixes. PaReGTA-Date used the calendar date and appended the exact elapsed-day marker to subsequent visits; PaReGTA-Gap used \texttt{first visit} for the initial visit and the number of days since the immediately preceding retained visit thereafter; and PaReGTA-Month expressed the same interval in 30-day month bins as \texttt{within N months after previous}.

Migraine diagnosis records were used to define the cohort and guide disease-specific record filtering; however, the migraine diagnosis codes, migraine subtype terms, and the \texttt{meta\_migraine} field used to derive the target were not serialized into the primary visit text or included in the predictor set. Fail-fast feature-contract checks rejected target, meta, migraine, label, outcome, and equivalent direct label-derived fields before model fitting. Raw patient identifiers and date fields were used for record alignment, chronological ordering, temporal-feature construction, and sequence truncation rather than supplied directly as predictor variables. Downstream PaReGTA classifiers received demographic variables and learned medication and comorbidity representations, whereas the patient-level target was supplied separately.

\begin{table}[t]
\tbl{Simplified illustration of PaReGTA-Gap visit text derived from the constructed EHR example in Tables~\ref{table_example_data} and~\ref{table_example_data2}.}
{\begin{tabular}{cccc} 
\toprule
ID & Visit date & Concept & Text \\
\midrule
1 & 2021-06-01 & Medication & first visit, meds: acetaminophen 325 MG Oral Tablet. \\
1 & 2021-06-01 & Comorbidities & first visit, comorbidities: Depression. \\
\midrule
1 & 2021-09-01 & Medication & \makecell{92 days after previous, \\ meds: lasmiditan 100 MG Oral Tablet.} \\
1 & 2021-09-01 & Comorbidities & \makecell{92 days after previous, \\ comorbidities: Depression ; Insomnia.} \\
\midrule
1 & 2021-12-01 & Medication & \makecell{91 days after previous, \\ meds: lasmiditan 100 MG Oral Tablet ; ibuprofen.} \\
1 & 2021-12-01 & Comorbidities & \makecell{91 days after previous, \\ comorbidities: Depression ; Insomnia.} \\
\bottomrule
\end{tabular}}
\label{table_visit_text_example}
\end{table}

Table~\ref{table_visit_text_example} provides a simplified illustration of the PaReGTA-Gap visit-text structure derived from the preprocessed example in Table~\ref{table_example_data2}. Because the constructed source records do not specify the medication-class labels appended by the implementation, the displayed medication strings are schematic rather than byte-exact encoder inputs; the temporal prefixes, semicolon delimiters, and terminal punctuation follow the implemented text builder.

As the base sentence-embedding model, we used GTE-base-v1.5 (\texttt{Alibaba-NLP/gte-base-en-v1.5}) from the GTE model family \citep{zhang-etal-2024-mgte}. We selected this model because it generates fixed-dimensional text embeddings, was computationally feasible for fine-tuning and large-scale inference in our environment, and integrates with the Sentence-Transformers workflow used in PaReGTA. The Massive Text Embedding Benchmark (MTEB) provides a standardized multi-task benchmark for evaluating general-purpose text embeddings across diverse task categories \citep{muennighoff-etal-2023-mteb}.

\subsubsection{Domain adaptation through SimCSE-style contrastive fine-tuning}

To adapt the base encoder to cohort-specific visit texts, we performed SimCSE-style identical-text contrastive fine-tuning with GTE-base-v1.5 (\texttt{Alibaba-NLP/gte-base-en-v1.5}) using the Sentence-Transformers \texttt{MultipleNegativesRankingLoss} implementation \citep{reimers_sbert_2019,gao_simcse_2021}. For each retained visit text $x_i$, we constructed an identical input pair $(x_i,x_i)$, and the model encoded the two copies in separate stochastic forward passes. Let $u_i=f_{\theta}(x_i)$ and $v_i=f_{\theta}(x_i)$ denote the embeddings from the first and second elements of pair $i$. For a minibatch of $B$ pairs, the implemented objective was

\begin{equation}
\displaystyle
\mathcal{L}_{\mathrm{MNR}}
=
-\frac{1}{B}\sum_{i=1}^{B}
\log
\frac{\exp\!\left(s\,\mathrm{sim}(u_i,v_i)\right)}
{\sum_{j=1}^{B}\exp\!\left(s\,\mathrm{sim}(u_i,v_j)\right)}.
\end{equation}

Here, $\mathrm{sim}(u,v)=u^{\top}v/(\lVert u\rVert\lVert v\rVert)$. Because the loss was instantiated without overriding its similarity function or scale, it used cosine similarity and the default scale $s=20$, equivalent to a temperature of $1/s=0.05$. For each first-view embedding $u_i$, its paired second-view embedding $v_i$ was treated as the positive, whereas the second-view embeddings $\{v_j:j\neq i\}$ from the other batch examples were treated as in-batch negatives. The implemented objective optimized only the first-to-second-view direction and did not include a reverse-direction term. Thus, it encouraged separate stochastic embeddings of the same visit text to remain close while ranking the paired copy above the other second-view batch examples; it did not establish that those other texts were clinically unrelated.

We fit the encoder using only nonempty visit texts from the canonical outer-training PID set. Before fine-tuning, we removed exact-duplicate strings, shuffled the remaining unique texts, and retained at most 200,000 texts. Visit texts for test and feature importance were not used to fit the encoder. We trained a separate domain-adapted encoder for each concept configuration. This sentence-encoder adaptation did not use the patient-level target. In contrast, the attention pooler used in the subsequent patient-level aggregation step was trained with patient triplets constructed from target values, as described below. All subsequent experiments used the Visit-text + SimCSE setting as the default visit-level encoder.

\subsubsection{Encoding visit-level representations}

After textualization, the fine-tuned sentence-embedding model mapped the completed medication and comorbidity strings to visit-level vectors; it did not generate, reorder, or impute EHR content. In other words, rather than encoding all textualized visit data of a patient directly into a patient-level representation, we adopted a hierarchical strategy, in which low-level units (visits) are first embedded and then pooled at the patient level. This is because prior work on hierarchical document modeling has shown that encoding sentences or paragraphs separately and aggregating them through a higher-level mechanism often yields better performance and interpretability than encoding the entire document in one pass \citep{yang_han_2016}. Similar hierarchical strategies have been applied to longitudinal clinical text, where sentences are aggregated into note-level representations and multiple notes are subsequently combined into a patient-level representation \citep{SiRoberts2020PatientRepresentation}. This hierarchical design respects the natural structure of EHR data (patients, visits, and events) and allows pooling mechanisms to modulate the relative importance of different visits.

Unlike the sparse baselines, which represented higher-level medication concepts as patient-level indicators or counts, PaReGTA encoded the deterministically constructed product-level medication strings at the visit level. This design reduced additional task-specific aggregation into coarse features, but it still required cohort construction, retained standardized concepts, and medication preprocessing.

\subsubsection{Hybrid temporal pooling into patient representations}

To obtain patient-level representations, we combined two complementary pooling strategies over visit embeddings: a time-weighted scheme and an attention-based scheme. The time-weighted component assigns larger weights to more recent visits, reflecting the intuition that recent clinical events are often more relevant to current risk. The attention component identifies visits that are globally important for prediction, regardless of their recency, and amplifies their contribution. Let $\mathbf{v}_i$ denote the embedding of visit $i$, let $t_i$ denote its date, and let $t_N$ denote the date of the patient's most recent retained visit. For the reported analyses, we first calculated the nonnegative elapsed time in days and then applied the configured \texttt{log1p} transformation:
\begin{equation}
\begin{aligned}
d_i &= \max\!\left\{(t_N-t_i)_{\mathrm{days}},0\right\}, \\
\delta_i &= \log(1+d_i).
\end{aligned}
\end{equation}
The unnormalized recency score and normalized time-weighted coefficient were then defined as
\begin{equation}
r_i=\exp(-\gamma\delta_i)=(1+d_i)^{-\gamma}, \qquad
w_i^{\mathrm{time-weighted}}
=\frac{r_i}{\sum_{j=1}^{N}r_j}.
\end{equation}

Here, $d_i$ represents the elapsed days from visit $i$ to the most recent retained visit, rather than the interval between consecutive visits used in Gap textualization. The fixed parameter $\gamma>0$ controls the strength of decay on the log-transformed elapsed-time scale, with larger values assigning relatively less weight to older visits. It should therefore not be interpreted as a raw per-day exponential decay rate.

For the attention component, we concatenated each visit embedding $\mathbf{v}_i \in \mathbb{R}^{D}$ with the same log-transformed elapsed-time feature $\delta_i$ defined above and passed the resulting vector through a two-layer multilayer perceptron:
\begin{equation}
\tilde{s}_i = \mathrm{MLP}\bigl([\mathbf{v}_i;\, \delta_i]\bigr), \qquad \mathrm{MLP}: \mathbb{R}^{D+1} \to \mathbb{R},
\end{equation}
where the MLP consists of a linear layer mapping to a hidden dimension $H$, a ReLU activation, and a second linear layer projecting to a scalar. The attention weights are then obtained via softmax:
\begin{equation}
w_i^{\text{attention}} = \frac{\exp(\tilde{s}_i)}{\sum_{j=1}^{N} \exp(\tilde{s}_j)}.
\end{equation}
This formulation allows the model to learn flexible, nonlinear interactions between visit content and temporal position, rather than relying on a fixed functional form. By including the time gap as an explicit input, the attention mechanism can jointly consider both what was recorded and when it was recorded. 

The sentence encoder and the attention pooler used different forms of supervision. We adapted the sentence encoder using the unsupervised SimCSE objective, whereas we trained the attention pooler using the binary patient-level target. Attention-pooler training was restricted to patients present in both the designated training visit-embedding data and the corresponding training-label file; held-out test and attribution labels were not used. For each reference patient, we sampled up to four distinct positive patients with the same target value and one negative patient with the other target value for each reference-patient--positive-patient pair.

Let $\mathbf{z}_{\mathrm{time}}=\sum_{i=1}^{N}w_i^{\mathrm{time\mbox{-}weighted}}\mathbf{v}_i$ and $\mathbf{z}_{\mathrm{attn}}=\sum_{i=1}^{N}w_i^{\mathrm{attention}}\mathbf{v}_i$ denote the time-weighted and attention-pooled representations, respectively. We define the implemented hybrid patient representation as
\begin{equation}
    \mathbf{z}
    =
    \alpha\mathbf{z}_{\mathrm{attn}}
    +
    (1-\alpha)\mathbf{z}_{\mathrm{time}},
\end{equation}
where $\alpha\in[0,1]$ denotes the contribution of the attention-pooled representation and $1-\alpha$ denotes the contribution of the time-weighted representation. Thus, $\alpha=1$ corresponds to attention-only pooling, whereas $\alpha=0$ corresponds to time-weighted-only pooling. The implementation does not apply an additional L2-normalization step after this convex combination. For more details on the hyperparameters of PaReGTA, please refer to Appendix Subsections~\ref{subsec:detail_hyperparameter} and~\ref{subsec:hyperparameter_experiment}.

This hybrid pooling strategy provides a flexible way to encode both the recency of visits and their global importance into a single patient-level vector that can be used by standard machine learning models. The resulting patient-level representation is also used in the PaReGTA-RSS factor-removal sensitivity analysis described in Subsection~\ref{subsection_rss}.

\subsection{Downstream model development and evaluation}

After constructing patient-level representations, feature preprocessing, downstream model development, and held-out evaluation followed the study-specific experimental protocol. For the candidate analyses reported in the main text, representation learning, fitted preprocessing, and classifier development were restricted to the designated training data. We used a fixed outer patient-level partition that separated the training pool, held-out test cohort, and attribution cohort.

Fitted preprocessing transformations were learned from the designated training data and applied to held-out data without refitting. When an inner validation subset was used, model and configuration selection relied only on validation performance, and the fixed test cohort was reserved for final evaluation. Factor-removal analyses were conducted using the separately designated attribution data.

We used patient-level bootstrap resampling of fixed held-out test predictions to quantify sampling uncertainty without retraining models or repeating the data partitions; detailed methods are provided in Appendix~\ref{subsec:fixed_test_bootstrap}.

To examine sensitivity to documentation intensity, we performed two exploratory analyses based on the number of unique recorded dates: a post-hoc subgroup evaluation of fixed-test predictions from PaReGTA-Gap + LightGBM developed without visit-count stratification and cohort-specific retraining of PaReGTA-Gap, T-LSTM, and RETAIN within the same visit-count strata. Detailed procedures are provided in Appendix~\ref{subsec:visit_count_analysis}.

\subsection{PaReGTA-RSS factor-removal sensitivity analysis}
\label{subsection_rss}

We used PaReGTA-RSS (Representation Shift Score) as a prespecified factor-removal sensitivity analysis to quantify how a fitted model's logit changed after a clinically defined factor group was removed from the observed EHR input. For each patient $i$ and prespecified factor group $I$, such as a medication or comorbidity factor group, we removed all occurrences of $I$ from the visit-level input and recomputed the complete PaReGTA text$\rightarrow$embedding$\rightarrow$pooling pipeline. We applied the same fitted PyCaret preprocessing transformation to the original and factor-removed representations and kept the sentence-embedding model, pooling parameters, preprocessing transformation, and downstream logistic-regression model fixed. No model or preprocessing component was retrained or refit after factor removal.

Let $\mathbf{x}_i$ and $\mathbf{x}_i^{(-I)}$ denote the transformed original and factor-removed representations after applying the fitted preprocessing transformation, respectively. Let $\mathbf{c}$ and $b$ denote the coefficient vector and intercept of the fixed logistic-regression model. We define the signed patient-level RSS as
\begin{equation}
\label{eq_score_lr}
\begin{split}
\mathrm{RSS}_{i}(I)
&=
\left(\mathbf{c}^{\top}\mathbf{x}_i+b\right)
-\left(\mathbf{c}^{\top}\mathbf{x}_i^{(-I)}+b\right)\\
&=
\mathbf{c}^{\top}\left(\mathbf{x}_i-\mathbf{x}_i^{(-I)}\right).
\end{split}
\end{equation}
Because the intercept cancels, $\mathrm{RSS}_{i}(I)$ is exactly the change in the fixed logistic-regression logit under the specified factor-removal operation. The sign indicates the direction of the fitted-logit change, whereas the absolute value indicates its magnitude. For medication and comorbidity summaries, we reported the cohort-level magnitude as the mean of $|\mathrm{RSS}_{i}(I)|$ across the patients included in the corresponding analysis. For the temporal analysis, we defined four nested factor groups as the retained medication and comorbidity records within 30, 90, 180, or 365 days before each patient's most recent retained visit. To keep cohort composition constant across horizons, we used the patients eligible for the 365-day analysis as a fixed common cohort and required the same patients to be present in the original and all four factor-removed representations. For target group \(g\), temporal sensitivity was summarized as the mean absolute fitted-logit change, \(\frac{1}{n_g}\sum_{i\in g}|\mathrm{RSS}_{i}(T)|\).

PaReGTA-RSS shares the general perturbation principle used by SHAP because both examine changes in model output after modifying the input. However, SHAP estimates coalition-weighted feature contributions to predictions \citep{lundberg_shap_2017}, whereas PaReGTA-RSS removes one prespecified factor group at a time, recomputes the complete patient representation, and reports the resulting change in the fixed fitted logit. PaReGTA-RSS is therefore not a SHAP estimator, a Shapley value, or a new general attribution theory. It should be interpreted as predictive sensitivity to the specified removal operation within this fitted pipeline, not as a clinical association, causal or treatment effect, or evidence of biological importance.

\section{Experiments and Results}

\subsection{Migraine case-study cohort and analytical task}

We evaluated PaReGTA using a retrospective patient-level classification task with the binary EHR-derived target defined in Subsection~\ref{subsec:patient_level_target}. The analytic cohort included 39,088 patients: 7,609 (19.47\%) were target-positive under the ever-chronic rule, whereas 31,479 (80.53\%) were target-negative. These target groups were mutually exclusive and exhaustive.

We grouped migraine diagnosis records into six operational subtype categories: migraine without aura, migraine with aura, chronic migraine without aura, menstrual migraine, other migraine, and unspecified migraine. The codebook additionally included ICD-9-CM 346 and ICD-10-CM G43 as broad migraine parent codes, which were listed separately from the subtype categories. A patient could receive records from more than one category over time. The corresponding nonexclusive patient counts are reported in Table~\ref{table_migraine_subtype_distribution}.

\begin{table}[!h]
\tbl{Patients with at least one recorded code in each nonexclusive operational migraine subtype category.}
{\begin{tabular}{lcc}
\toprule
Operational subtype & Patients, n & Cohort proportion, \% \\
\midrule
Migraine without aura & 13,053 & 33.39 \\
Migraine with aura & 10,799 & 27.63 \\
Chronic migraine without aura & 7,609 & 19.47 \\
Menstrual migraine & 694 & 1.78 \\
Other migraine & 5,796 & 14.83 \\
Unspecified migraine & 30,036 & 76.84 \\
\bottomrule
\end{tabular}}
\label{table_migraine_subtype_distribution}
\end{table}

Subtype categories were nonexclusive because a patient could have records from more than one category during the observation period. Each percentage was calculated using the full cohort of 39,088 patients as the denominator; therefore, the row counts and percentages should not be summed.

The longitudinal code-history audit classified 31,479 patients (80.53\%) as non-chronic-only, 828 (2.12\%) as chronic-only, 934 (2.39\%) as having a one-way non-chronic-to-chronic code history, 498 (1.27\%) as having a one-way chronic-to-non-chronic code history, and 5,349 (13.68\%) as having a complex mixed history. The complex category combined different patterns---same-day chronic/non-chronic co-recording, repeated changes in direction, and mixed histories whose order could not be determined because dates were missing---and should not be interpreted as a single type of temporal conflict. All 5,349 patients in the observed complex category had at least one qualifying chronic record; therefore, the chronic-only, two one-way-transition, and complex categories together comprised the 7,609 target-positive patients under the ever-chronic rule.

For the primary PaReGTA analysis, we used demographic variables together with learned representations of longitudinal medication and comorbidity records as predictor inputs; the patient-level target was supplied separately. Encounter metadata includes hospital, clinic, and emergency-room care records. Comorbidity profiling shows frequent cardiometabolic and mental health conditions: hypertension is highly prevalent, and depression, insomnia, and anxiety are common. We utilized data from 14 migraine-relevant medication classes (acetaminophen, antiemetics, beta-blockers, botulinum, CGRP, ergot, gabapentin, gepant, lasmiditan, NSAID, opioids, tricyclic, triptan, and verapamil). The demographic variables comprised age, sex at birth, and race. Age was calculated based on the date of birth and the date of the latest visit record. After cohort construction and preprocessing, the analytic cohort included 39,088 patients with at least one recorded migraine diagnosis. Overall, 80\% of the patients were assigned female at birth. By race, 61\% were White and 10\% were Black. Most patients were adults aged 18--64 years (81\%), and 14\% were older adults (aged $\geq 65$ years). Visit counts and observation spans were right-skewed. Across the full analytic cohort, the median number of visits was 3, with a mean of 15.45 visits (SD, 42.66; range, 1--1,657). The corresponding medians were 2 visits among the 31,479 target-negative patients and 14 visits among the 7,609 target-positive patients. The corresponding means were 12.04 visits (SD, 39.84; range, 1--1,657) and 29.56 visits (SD, 50.37; range, 1--850), respectively.

The median observation span was 19.65 months across the full analytic cohort, 10.15 months in the target-negative group, and 57.43 months in the target-positive group. The corresponding means were 45.40 months (SD, 59.41; range, 0--475.60), 39.56 months (SD, 57.28; range, 0--475.60), and 69.58 months (SD, 61.89; range, 0--441.10), respectively. The longer observed histories in the target-positive group may reflect greater opportunity for qualifying chronic migraine codes to be recorded and should not be interpreted as evidence of greater biological severity or a causal effect of visit frequency.

\subsection{Experimental setup}

For experiments with the proposed framework, we followed the full pipeline described in the previous section: textualization, target-free SimCSE adaptation and label-supervised attention-pooler training, encoding visit texts into visit-level embedding vectors, and aggregating the vectors into patient-level representations (see Figure~\ref{fig_llm_pipeline}). Table~\ref{table_visit_text_example} shows the textualized version of the example data in Table~\ref{table_example_data2}, which serves as input to the sentence-embedding LLM. After constructing patient-level representations via PaReGTA, we standardized all features to zero mean and unit variance. We then applied principal component analysis (PCA) for dimensionality reduction, retaining the minimum number of components that explained 95\% of the total variance for each concept-specific representation (original dimensionality: 768). We report classification performance on the test set and use the attribution set exclusively for factor-importance analyses.

We used PyCaret's classification module to train and compare machine-learning models on the processed representations. For the main representation-comparison experiments implemented in PyCaret, we used the same classifier setup for each encoded dataset and enabled training-time imbalance correction through PyCaret's \texttt{fix\_imbalance} option. This procedure was not applied to MedBERT, CEHR-BERT, T-LSTM, or RETAIN, which followed their respective training protocols. We focused on five representative classifiers implemented in PyCaret: LightGBM, Gradient Boosting Classifier (GBC), Extreme Gradient Boosting (XGBoost), logistic regression (LR) with $\ell_2$ regularization, and a linear Support Vector Machine (SVM). For each encoded dataset, we fitted each of the five configured classifier families using the same PyCaret setup and generated one fixed-test prediction file for each representation--classifier combination. The archived workflow did not call an additional hyperparameter-tuning or classifier-comparison routine, and no validation-only artifact was preserved for selecting a single representation--classifier pair. For details on the experimental environment and settings, see Appendix \ref{app_environment} and \ref{app_ex_setting}. 

To characterize the performance change under deliberate direct label inclusion, we constructed an intentional positive control by adding the label-derived migraine-type field to an otherwise parallel PaReGTA-Gap input. We regenerated the corresponding patient representations, retrained LightGBM, and evaluated the model on the same fixed held-out test cohort of 7,818 patients. We treated this configuration only as a positive control and did not include it among the candidate representations.

We evaluated PaReGTA-Gap against sparse baselines, LLM-based models (MedBERT and CEHR-BERT), and time-series-based models (T-LSTM and RETAIN) using the same outer patient-level partition. We generated this partition once by deterministic contiguous slicing, without shuffling, target stratification, or a random seed: 27,361 patients (70\%) formed the outer training pool, 7,818 patients (20\%) formed the fixed held-out test cohort, and 3,909 patients (10\%) were reserved for attribution analyses. For the deep-learning comparators, we further divided the frozen outer training pool into 24,624 train-fit and 2,737 validation patients using a target-stratified 90/10 split with seed 0. This inner split did not alter the fixed test or attribution cohorts. For these deep-learning comparators, model and configuration selection used only the validation subset, and the fixed test cohort was used only for final evaluation. For more details on the experimental settings of the baseline models, please refer to Subsection~\ref{subsec:details_baseline}. 

\subsection{Performance comparison between data encoding methods}

\begin{table}[t]
\tbl{Performance comparison of models trained on encoded data with medication information. We underlined and bolded the best result and used bold for the second-best result among all results.}
{
\begin{tabular}{lcccccc} 
\toprule
Encoding Type & \multicolumn{3}{c}{Descriptions} & \multicolumn{3}{c}{\makecell{Examples in text\\ (previous visit: 2021-06-01)}} \\
\midrule
One-hot encoding & \multicolumn{6}{l}{Binary presence/absence indicators aggregated at the patient level.} \\
Count BoC & \multicolumn{6}{l}{Patient-level occurrence counts without temporality or ordering.}  \\
\midrule
PaReGTA-Date & \multicolumn{3}{l}{Calendar date} & \multicolumn{3}{l}{2022-01-01} \\
PaReGTA-Gap & \multicolumn{3}{l}{The date gap between neighboring visits.} & \multicolumn{3}{l}{214 days after previous} \\
PaReGTA-Month & \multicolumn{3}{l}{Summing up the date gap into months.} & \multicolumn{3}{l}{7 months after previous} \\
PaReGTA-Last & \multicolumn{3}{l}{Calculate the date gap from the last visit.} & \multicolumn{3}{l}{151 days before the latest visit} \\
PaReGTA-Without & \multicolumn{3}{l}{No temporal tokens.} & \multicolumn{3}{c}{-} \\
\midrule
\\
\midrule
Encoding Type & Metric & LightGBM & GBC & XGBoost & LR & SVM \\ 
\midrule
\multirow{2}{*}{One-hot encoding} & ACC & 84.02 & 83.78 & 83.38 & 80.35 & 80.10 \\
& AUC & 0.7634 & 0.7583 & 0.7549 & 0.762 & 0.7475 \\
\midrule
\multirow{2}{*}{Count BoC} & ACC & 84.73 & 84.15 & 83.96 & 82.6 & 83.17 \\
& AUC & 0.8354 & 0.8197 & 0.8272 & 0.7673 & 0.7256 \\
\midrule
\multirow{2}{*}{PaReGTA-Date} & ACC & 91.76 & 91.61 & 91.98 & 91.67 & 89.77 \\
& AUC & 0.9428 & 0.9416 & 0.9437 & 0.9453 & 0.9292 \\
\midrule
\multirow{2}{*}{PaReGTA-Gap} & ACC &
\textbf{92.29} &
92.18 &
\textbf{\underline{92.38}} &
91.57 &
91.14 \\
& AUC &
0.9501 &
\textbf{0.9514} &
0.9462 &
\textbf{\underline{0.9520}} &
-- \\
\midrule
\multirow{2}{*}{PaReGTA-Month} & ACC & 92.06 & 91.63 & 92.02 & 91.44 & 91.11 \\
& AUC & 0.9496 & 0.9481 & 0.9451 & 0.9495 & 0.9412 \\
\midrule
\multirow{2}{*}{PaReGTA-Last} & ACC & 79.42 & 80.29 & 82.17 & 79.68 & 77.76 \\
& AUC & 0.7493 & 0.7785 & 0.7659 & 0.7819 & 0.7362 \\
\midrule
\multirow{2}{*}{PaReGTA-Without} & ACC & 83.7 & 81.84 & 83.82 & 81.9 & 80.02 \\
& AUC & 0.8678 & 0.8636 & 0.8506 & 0.8636 & 0.8167 \\
\bottomrule
\end{tabular}}
\label{table_performance_data_preprocessing_with_m}
\end{table}

Using the representations and evaluation protocol defined in the Methods, we compared PaReGTA with the evaluated encoding and modeling approaches. To quantify how design choices in textualization of PaReGTA affect downstream performance, we performed an ablation study that varies (i) the representation of temporal information and (ii) whether medication information is explicitly included in the input text. 

We evaluated five temporal tokenization schemes in the ablation analysis: (i) \textit{Date}, using absolute-date tokens; (ii) \textit{Gap}, using the elapsed time between consecutive visits; (iii) \textit{Month}, using month-level discretization of the inter-visit gap; (iv) \textit{Last}, using the time from each visit to the most recent visit; and (v) \textit{Without}, which retained visit order without explicit temporal values. Because no contemporaneous prespecified or validation-only selection artifact was available, we treated PaReGTA-Gap + LightGBM as a post hoc analytical reference for the subsequent descriptive analyses. Tables~\ref{table_performance_data_preprocessing_with_m} and~\ref{table_performance_data_preprocessing_without_m} report the temporal ablation together with the one-hot and count-based BoC baselines.

\begin{table}[t]
\tbl{Temporal textualization ablation without medication. We underlined and bolded the best result and used bold for the second-best result among all results.}
{\begin{tabular}{lcccccc} 
\toprule
Encoding Type & \multicolumn{3}{c}{Descriptions} & \multicolumn{3}{c}{\makecell{Examples in text \\ (previous visit: 2021-06-01)}} \\
\midrule
PaReGTA-Gap & \multicolumn{3}{l}{The date gap between neighboring visits.} & \multicolumn{3}{l}{214 days after previous} \\
PaReGTA-Month & \multicolumn{3}{l}{Summing up the date gap into months.} & \multicolumn{3}{l}{7 months after previous} \\
PaReGTA-Last & \multicolumn{3}{l}{Calculate the date gap from the last visit.} & \multicolumn{3}{l}{151 days before the latest visit} \\
PaReGTA-Without & \multicolumn{3}{l}{No temporal tokens.} & \multicolumn{3}{c}{-} \\
\midrule
\\
\midrule
Encoding Type & Metric & LightGBM & GBC & XGBoost & LR & SVM \\ 
\midrule
PaReGTA-Gap & ACC & \textbf{\underline{92.06}} & 91.66 & \textbf{91.83} & 91.57 & 91.06 \\
& AUC & 0.9436 & \textbf{\underline{0.9456}} & 0.9398 & 0.9445 & 0.9405 \\
\midrule
PaReGTA-Month & ACC & 91.66 & 91.21 & 91.26 & 90.94 & 90.88 \\
& AUC & \textbf{0.9447} & 0.9444 & 0.9411 & 0.9406 & 0.9327 \\
\midrule
PaReGTA-Last &  ACC & 74.8 & 76.64 & 77.81 & 74.07 & 72.92 \\
& AUC & 0.6875 & 0.7303 & 0.7093 & 0.7246 & 0.6809 \\
\midrule
PaReGTA-Without & ACC & 81.57 & 80.4 & 81.48 & 78.27 & 77.5 \\
& AUC & 0.8436 & 0.8424 & 0.8314 & 0.8324 & 0.7918 \\
\bottomrule
\end{tabular}}
\label{table_performance_data_preprocessing_without_m}
\end{table}

Overall, LLM-based embeddings with temporal information (\textit{Date} or \textit{Gap}) substantially outperform the baselines across all five classifiers. For instance, under LightGBM, accuracy increased from 84.02\% for one-hot encoding to 91.76\% for PaReGTA-Date and 92.29\% for PaReGTA-Gap, while AUC increased from 0.7634 to 0.9428 and 0.9501, respectively. In this fixed-test ablation, PaReGTA-Date and PaReGTA-Gap had higher point estimates than the one-hot and count-based BoC baselines across the five downstream classifiers. Gap showed the strongest overall point-estimate pattern among the examined temporal representations, although the leading classifier varied by metric.

These results suggest that a simple relative notion of recency (\textit{Gap}) is particularly effective for this task. The inter-visit gaps capture patient-specific progression pace more robustly than absolute calendar dates, which can vary widely across patients for reasons unrelated to migraine type. Removing calendar-date and elapsed-time values while retaining only ordinal visit order (\textit{Without}) led to a clear performance reduction relative to \textit{Date} and \textit{Gap}, particularly in AUC. This indicates that temporal context provides useful predictive signal in our setting, and that omitting it discards information that the downstream classifiers can leverage. Notably, the \textit{Last} encoding performed substantially worse than \textit{Gap}, with AUC dropping to 0.75–0.78. This approach represents each visit by its distance from the most recent visit rather than from the preceding visit, thereby disrupting the sequential structure and obscuring patterns in consecutive visit intervals. Under LightGBM, the \textit{Month} encoding yielded a slightly lower AUC than \textit{Gap} (0.9496 vs. 0.9501); this descriptive comparison did not isolate temporal granularity as the cause of the difference. These experimental results highlight that performance depends not only on whether temporal information is included, but also on how it is encoded. 

Visit-sequence lengths varied substantially across patients. We therefore used length-bucket batching for T-LSTM and RETAIN to group patients with similar sequence lengths within the same batch and address the resulting training instability. Both recurrent models completed training and evaluation on the fixed test cohort. T-LSTM achieved performance close to PaReGTA-Gap, whereas RETAIN showed lower AUC, accuracy, and F1.

Within this exploratory fixed-test evaluation, the PaReGTA-Gap + LightGBM reference had higher point estimates than the evaluated sparse, BERT-based, and recurrent alternatives in this cohort. Importantly, PaReGTA is flexible with respect to medication information, which addresses two common real-world constraints. When medication records were available, PaReGTA encoded deterministically constructed product-level strings rather than reducing them to the higher-level indicators used by the sparse baselines. Conversely, when medication records are missing, incomplete, or potentially confounded by post-diagnostic prescribing, PaReGTA still maintains strong performance using comorbidity-derived visit text alone. This flexibility reduces preprocessing burden while improving robustness across practical EHR settings.

\subsection{Performance comparison with baseline models}

\begin{table}
\tbl{Exploratory performance comparison of the post hoc PaReGTA-Gap + LightGBM analytical reference and the evaluated EHR representation and sequence-model alternatives on the fixed test cohort}
{
\begin{tabular}{lccc}
\toprule
Model &
AUC &
accuracy &
F1 \\
\midrule
PaReGTA-Gap + LightGBM &
0.9501 &
0.9229 &
0.8147 \\
One-hot encoding + LightGBM &
0.7634 &
0.8402 &
0.4581 \\
BoC + LightGBM &
0.8354 &
0.8473 &
0.5162 \\
MedBERT &
0.8600 &
0.7970 &
0.5831 \\
CEHR-BERT &
0.9173 &
0.8175 &
0.6566 \\
T-LSTM &
$0.9447 \pm 0.0011$ &
$0.9093 \pm 0.0042$ &
$0.7975 \pm 0.0053$ \\
RETAIN &
$0.8559 \pm 0.0024$ &
$0.7632 \pm 0.0114$ &
$0.5544 \pm 0.0078$ \\
\bottomrule
\end{tabular}
}
\label{table_master_comparator}
\end{table}

We conducted comparative experiments with one-hot encoding, BoC, MedBERT, CEHR-BERT, T-LSTM, and RETAIN on the same fixed held-out test cohort. The following table summarizes their performance relative to PaReGTA-Gap. For the detailed explanation of the base models, please refer to Subsection~\ref{subsec:details_baseline}.

Table~\ref{table_master_comparator} compares sparse, BERT-based, and recurrent representations on the same fixed held-out test cohort. PaReGTA-Gap achieved the highest AUC, accuracy, and F1 among the evaluated methods, while T-LSTM provided the closest comparator across all three metrics. CEHR-BERT achieved higher performance than MedBERT. These findings are specific to the present cohort and evaluation protocol and do not establish the general superiority of one representation across datasets or clinical tasks.

PaReGTA-Gap, the sparse baselines, MedBERT, and CEHR-BERT are reported as fixed-test point estimates, whereas T-LSTM and RETAIN are summarized as the mean $\pm$ sample standard deviation across five training seeds.

In the fixed-prediction bootstrap analysis, the paired differences between PaReGTA-Gap + LightGBM and T-LSTM were positive for AUC, accuracy, and F1, and all three 95\% confidence intervals excluded zero (Appendix Tables~\ref{table_fixed_test_bootstrap_ci} and~\ref{table_paired_bootstrap_tlstm}).

In both visit-count designs, PaReGTA-Gap + LightGBM had higher AUC and F1 but lower accuracy in the stratum with at least 8 visits than in the 1--7-visit stratum. In the cohort-specific analysis, it had the highest reported AUC, accuracy, and F1 among PaReGTA-Gap, T-LSTM, and RETAIN within each stratum (Appendix Tables~\ref{table_visit_count_posthoc} and~\ref{table_visit_count_retraining}).

\subsection{Verification of embedding by LLM}

We examined whether PaReGTA yields a well-behaved representation space under different encoding strategies. We report metrics that are widely used to assess the geometric quality of sentence embeddings. Uniformity follows \citet{wang2020understanding} and measures how evenly embeddings are spread over the unit hypersphere; more negative values indicate that pairwise distances are more uniformly distributed and that the representation is less collapsed. Spectral flatness summarizes the spread of singular values of the embedding covariance matrix; higher values indicate that variance is distributed more evenly across dimensions rather than concentrated in a few directions. Finally, Top1 denotes the fraction of total variance explained by the first principal component; lower values indicate that the representation is not dominated by a single direction of maximal variance.

\begin{table}[t]
\tbl{Verification of PaReGTA representation quality by sentence embedding}
{\begin{tabular}{lccc} 
\toprule
\multirow{2}{*}{Method} &
\multirow{2}{*}{Uniformity ($\downarrow$)} &
\multicolumn{2}{c}{Isotropy} \\
& & Spectral flatness ($\uparrow$) & Top1 ($\downarrow$) \\
\midrule
Full-text & -0.1954 & 0.0217 & 0.2930 \\   
Visit-text & -1.3571 & 0.0254 & 0.2247 \\   
Visit-text + SimCSE & \textbf{-3.1857} & \textbf{0.0476} & \textbf{0.1204} \\  
\bottomrule
\end{tabular}}
\label{table_embedding_verification}
\end{table}

Table~\ref{table_embedding_verification} compares three settings: (i) \emph{Full-text}, where each patient’s longitudinal record is concatenated into a single long document, (ii) \emph{Visit-text}, where records are segmented into visit-level sentences without any further tuning, and (iii) \emph{Visit-text + SimCSE}, where the visit-level encoder is further fine-tuned using SimCSE on the dataset. As shown in the table, both visit-level encoding and SimCSE fine-tuning substantially improve these geometric properties compared to other settings. Taken together, these results indicate that our final configuration (Visit-text + SimCSE) yields an embedding space that is more uniform and isotropic than naive full-text or untuned visit-level representations. Although we do not explicitly model fine-grained temporal dynamics in this work, this analysis supports our design choice to (i) represent EHR information at the visit level and (ii) apply lightweight contrastive fine-tuning. 

\subsection{Ablation Study on fine-tuning LLM and training pooling}

\begin{table}[t]
\tbl{Ablation study on PaReGTA. We underlined and bolded the best result and used bold for the second-best result among all results.}
{\begin{tabular}{cccccccc} 
\toprule
SimCSE & Attention pooling & Metric & LightGBM & GBC & XGBoost & LR & SVM \\
\midrule
\multicolumn{2}{c}{\multirow{2}{*}{Original method}} & ACC & \textbf{\underline{92.16}} & 91.66 & \textbf{91.83} & 91.57 & 91.06 \\
& & AUC & 0.9439 & \textbf{\underline{0.9456}} & 0.9398 & 0.9445 & 0.9405 \\
\midrule
\multirow{2}{*}{N} & \multirow{2}{*}{N} & ACC & 91.16 & 90.93 & 91.23 & 91.62 & 90.98 \\
& & AUC & 0.9397 & 0.9374 & 0.9332 & 0.9379 & 0.9301 \\
\midrule
\multirow{2}{*}{N} & \multirow{2}{*}{Y} & ACC & 90.25 & 90.0 & 90.04 & 91.74 & 91.26 \\
& & AUC & 0.9095 & 0.9255 & 0.9135 & 0.9441 & 0.937 \\
\midrule
\multirow{2}{*}{Y} & \multirow{2}{*}{N} & ACC & 90.56 & 90.45 & 90.91 & 91.49 & 90.82 \\
& & AUC & 0.9247 & 0.9331 & 0.925 & 0.9368 & 0.9303 \\
\midrule
\multicolumn{2}{c}{\multirow{2}{*}{Separated comorbidities}} & ACC & 92.01 & 91.79 & 91.85 & 91.63 & 91.19 \\
& & AUC & 0.9449 & 0.9431 & 0.9407 & \textbf{0.9454} & 0.9397 \\
\bottomrule
\end{tabular}}
\label{table_performance_ablation}
\end{table}

Table~\ref{table_performance_ablation} summarizes an ablation study that isolates the contribution of two core components in PaReGTA: SimCSE-based contrastive fine-tuning of the encoder and attention-based pooling for aggregating visit-level embeddings into a patient-level representation. The “Original method” uses both modules, while the other rows remove one or both modules to test whether performance gains depend on each component. Overall, removing either module degrades performance, and removing both leads to the largest drop, indicating that encoder adaptation and informed aggregation are complementary. In particular, disabling SimCSE while keeping attention pooling results in a substantial reduction in AUC for several models, while enabling SimCSE without attention pooling yields intermediate performance. A structural variant that separates comorbidities before embedding shows performance very close to the original method, suggesting that the main improvements are driven by the two methodological modules rather than by minor formatting choices.

\subsection{PaReGTA-RSS factor-removal sensitivity results}

We applied PaReGTA-RSS using the fitted logistic-regression model to prespecified medication and comorbidity factor groups. For each patient and factor group, we removed the specified factor from the visit-level text, recomputed the complete PaReGTA representation, applied the same fitted preprocessing transformation, and calculated the signed fitted-logit change according to Equation~\ref{eq_score_lr}. We summarized cohort-level sensitivity to medication and comorbidity removal using the mean absolute PaReGTA-RSS.

For visualization, we retained factors recorded in at least 100 patients in the full-cohort and female-sex-at-birth analyses and in at least 50 patients in the male-sex-at-birth analysis. These minimum-count thresholds determined which aggregate summaries were displayed; they did not alter the definition of the patient-level score.

\begin{figure}
\centering
\includegraphics[width=0.80\columnwidth]{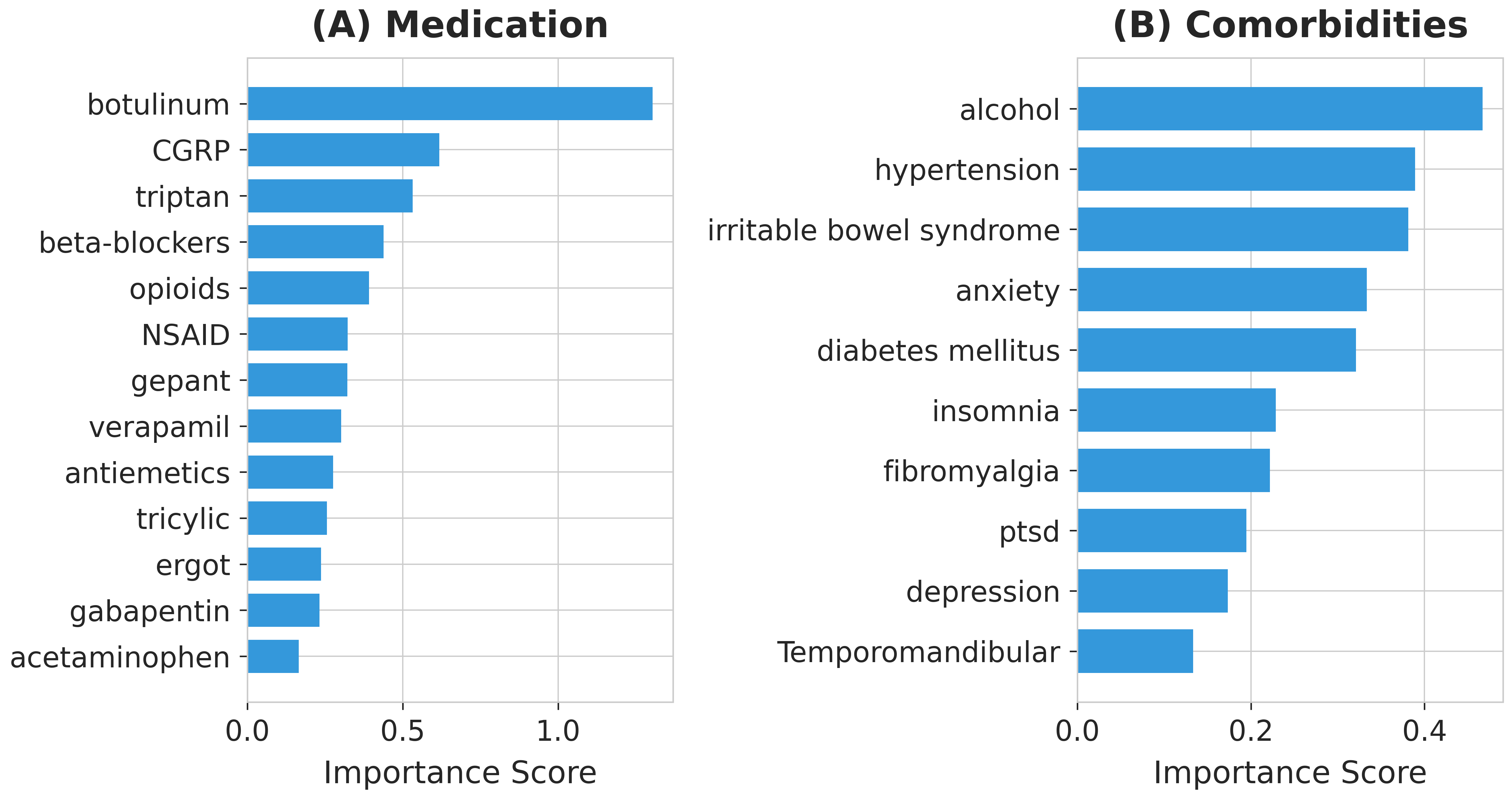}
\caption{PaReGTA-RSS factor-removal sensitivity for (A) medication classes and (B) comorbidities in the full cohort. The panels use independent horizontal scales.}
\label{fig_all_factor_importance}
\end{figure}

Figure~\ref{fig_all_factor_importance} presents the mean absolute PaReGTA-RSS for prespecified medication and comorbidity removals. Among the medication factor groups, removal of onabotulinumtoxinA produced the largest average absolute fitted-logit change, followed by removal of CGRP-targeting therapies and triptans. Among the comorbidity factor groups, removal of alcohol-related records, hypertension, irritable bowel syndrome, anxiety, and diabetes mellitus produced the largest average absolute changes. Anxiety had a relatively high mean absolute PaReGTA-RSS despite a near-zero mean signed score, indicating that its removal shifted the fitted logit in different directions across patients. These rankings identify the record removals to which this fitted pipeline was most sensitive; they do not estimate treatment effects, independent clinical associations, or biological importance.

\begin{figure}
\centering
\includegraphics[width=0.80\columnwidth]{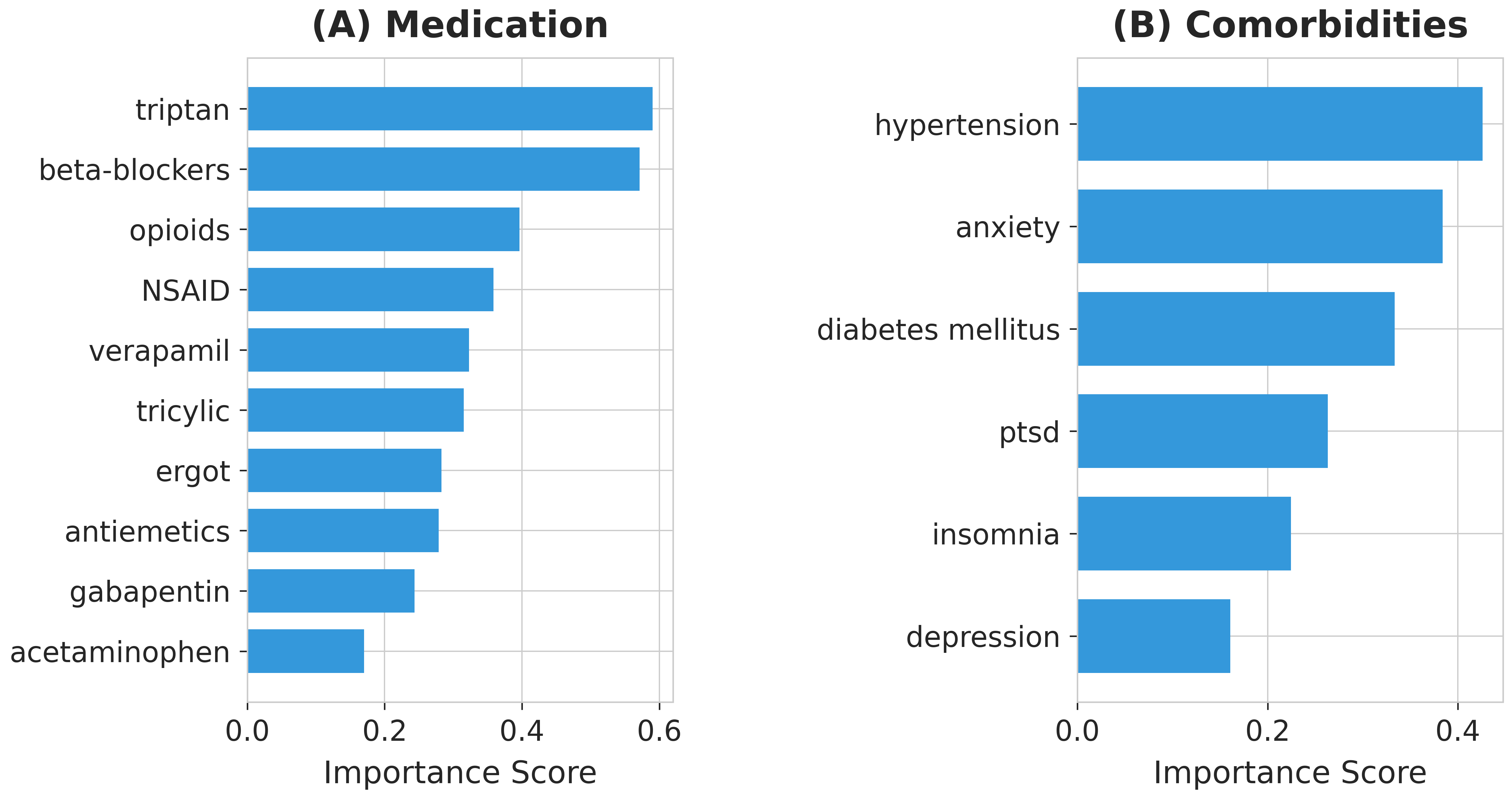}
\caption{PaReGTA-RSS factor-removal sensitivity for (A) medication classes and (B) comorbidities among participants recorded as male at birth. The panels use independent horizontal scales.}
\label{fig_male_factor_importance}
\end{figure}

\begin{figure}
\centering
\includegraphics[width=0.95\columnwidth]{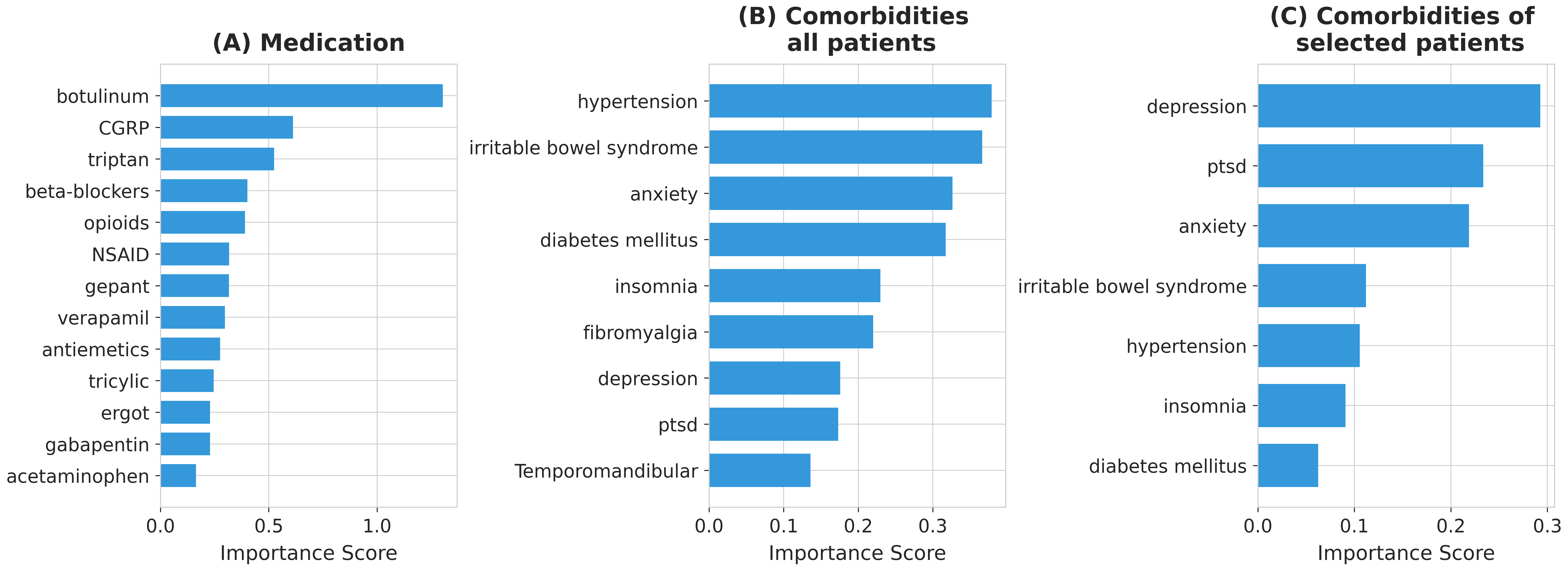}
\caption{PaReGTA-RSS factor-removal sensitivity among participants recorded as female at birth: (A) medication classes, (B) comorbidities in the full female subgroup, and (C) comorbidities in the subgroup without temporomandibular disorders or fibromyalgia. The panels use independent horizontal scales.}
\label{fig_female_factor_importance}
\end{figure}

After restricting the female subgroup to patients without temporomandibular disorders or fibromyalgia (Figures~\ref{fig_male_factor_importance} and \ref{fig_female_factor_importance}). Among participants recorded as male at birth, removal of triptan and beta-blocker records produced the largest mean absolute fitted-logit changes among the displayed medication groups, whereas removal of hypertension and anxiety records produced the largest changes among the displayed comorbidity groups. Among participants recorded as female at birth, removal of botulinum toxin records produced the largest change among the displayed medication groups, followed by removal of CGRP-targeting therapy and triptan records; removal of hypertension, irritable bowel syndrome, anxiety, and diabetes mellitus records produced the largest changes among the displayed comorbidity groups. 

After restricting the female subgroup to patients without temporomandibular disorders or fibromyalgia (Figure~\ref{fig_female_factor_importance}C), removal of depression and PTSD records produced the largest mean absolute fitted-logit changes among the displayed comorbidity groups. This subgroup analysis describes variation in the fitted pipeline's sensitivity to the prespecified record-removal operations and does not establish clinical importance, causal effects, or individualized diagnostic implications.

\begin{figure}
\centering
\includegraphics[width=0.98\columnwidth]{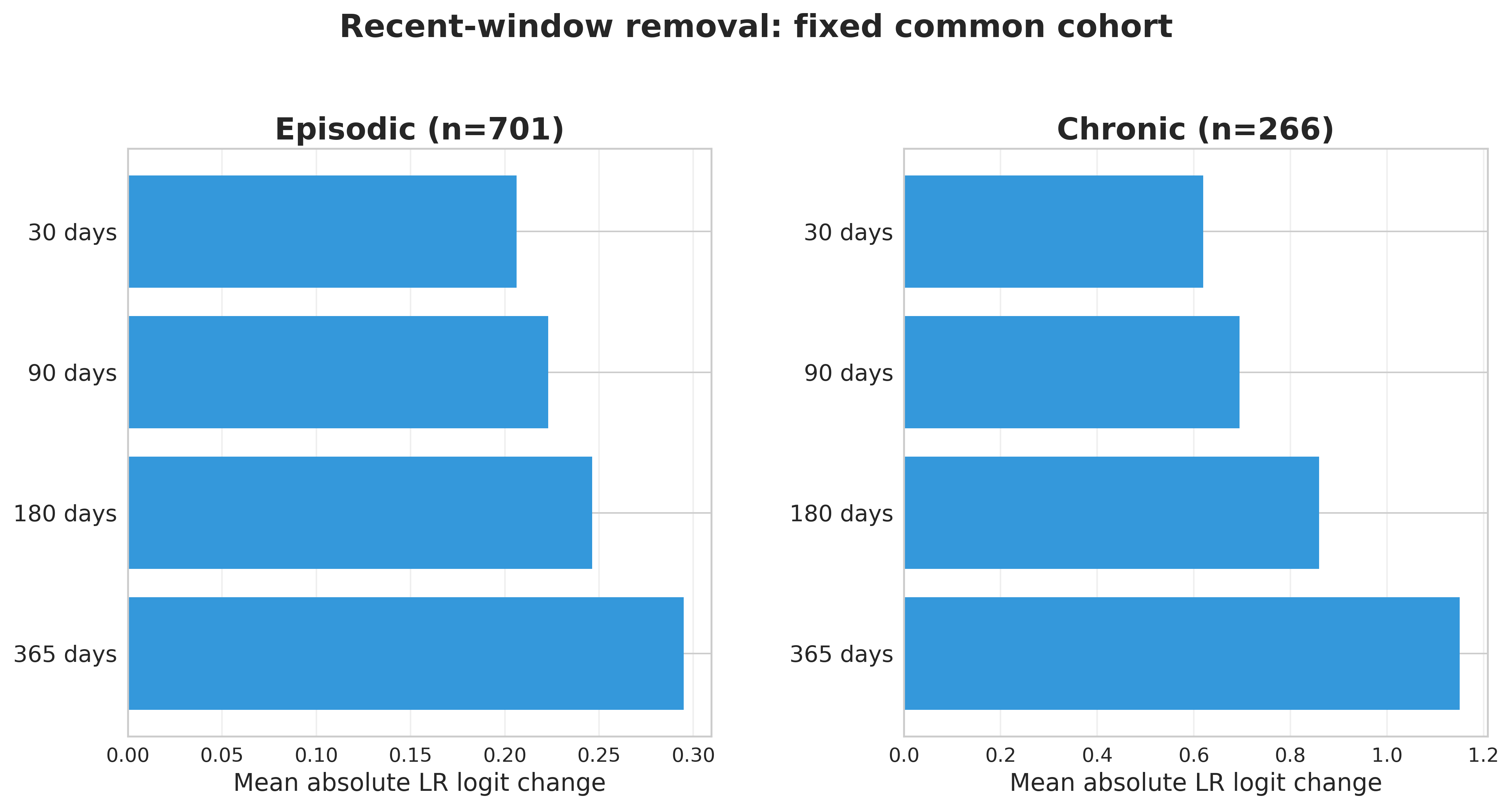}
\caption{Temporal PaReGTA-RSS sensitivity after removing retained medication and comorbidity records within nested 30-, 90-, 180-, and 365-day recency windows in a fixed common cohort: (A) the target-negative group (\(n=701\)) and (B) the target-positive group (\(n=266\)). Values represent mean absolute fitted-logit changes, and the panels use independent horizontal scales.}
\label{fig_back_factor_importance}
\end{figure}

We extended PaReGTA-RSS to temporal factors by removing retained medication and comorbidity records within nested 30-, 90-, 180-, and 365-day windows before each patient's most recent retained visit. To permit direct comparisons across horizons, we evaluated all four removal windows in a fixed common cohort of 967 patients, including 701 target-negative and 266 target-positive patients. Figure~\ref{fig_back_factor_importance} reports the mean absolute fitted-logit change within each target group.

The mean absolute fitted-logit change increased from 0.2062 at 30 days to 0.2949 at 365 days in the target-negative group and from 0.6192 to 1.1510 in the target-positive group. The values were consistently larger in the target-positive group, indicating that the fitted classifier was more sensitive to removal of recent longitudinal records in this group within the selected long-observation cohort. Because the recency windows were nested, this pattern represents cumulative sensitivity to removing progressively broader portions of recent recorded history rather than the independent contribution of discrete time intervals or a per-day effect. These results do not establish the clinical or causal importance of a particular time window.

\section{Conclusion and Discussion}\label{s:3:discussion}

This study evaluated PaReGTA as a reproducible patient-level framework for structured longitudinal EHR analysis and examined how temporal representation choices influenced performance in the migraine case study.

In the exploratory fixed-test comparison, the post hoc PaReGTA-Gap + LightGBM analytical reference had higher reported AUC, accuracy, and F1 than the evaluated sparse, BERT-based, and recurrent alternatives, while T-LSTM provided the closest comparator. The paired fixed-prediction bootstrap differences between PaReGTA-Gap + LightGBM and T-LSTM were positive for all three metrics.

The temporal textualization analysis showed higher fixed-test point estimates for PaReGTA-Date and PaReGTA-Gap than for PaReGTA-Without across the five downstream classifiers. PaReGTA-Without omitted explicit temporal values from the visit text but retained the common temporal pooling procedure; therefore, this comparison supports the contribution of textualized temporal cues under the evaluated conditions rather than isolating the contribution of temporal information across the complete PaReGTA pipeline. By producing fixed-dimensional patient representations that can be used with conventional classifiers, PaReGTA provides a modular alternative to sparse aggregation and specialized sequence models.

PaReGTA-RSS complements the performance analysis by describing the prespecified factor removals to which the fitted pipeline was most sensitive. The variation across medication, comorbidity, sex-at-birth subgroup, and temporal factor-removal analyses indicates that fitted-logit sensitivity differed across the predefined factor groups and patient subsets examined in this cohort. These results do not estimate Shapley values, allocate independent effects among correlated or overlapping factor groups, or support causal, biological, or clinical interpretations. Accordingly, they should be interpreted as retrospective characterization of the fitted pipeline rather than guidance for independent clinical diagnosis.

This study has several limitations. First, PaReGTA-RSS depends on the selected factor definitions, removal rule, learned representation, fitted preprocessing transformation, and downstream logistic-regression model; different analysis choices may therefore yield different sensitivity values. Second, T-LSTM and RETAIN were trained using length-bucket batching, which grouped patients with similar sequence lengths to reduce padding inefficiency. Although the contrast with the preliminary unstable runs is consistent with sequence-length heterogeneity and padding inefficiency having contributed, we did not perform a controlled comparison of batching strategies and therefore cannot identify either factor as the sole cause. Third, while the experiments shown in Tables \ref{table_performance_data_preprocessing_with_m} and \ref{table_performance_data_preprocessing_without_m} demonstrate the usefulness of PaReGTA compared with one-hot encoding, the models in those experiments were trained using data that include periods after chronic migraine diagnosis, making them difficult to apply directly in real clinical settings. Accordingly, this study evaluates representation learning in a retrospective classification setting rather than prospective clinical risk prediction, and the primary purpose of these experiments is to validate the effectiveness of PaReGTA as a patient-representation method. 

All analyses were conducted in a single All of Us cohort and were limited to one retrospective patient-level classification task with an EHR-derived migraine target. Because we did not perform validation in an independent external EHR cohort, the extent to which the findings apply across institutions, populations, diseases, or analysis tasks remains untested. The patient-level bootstrap quantified sampling uncertainty within the fixed held-out test cohort; it did not capture distributional differences across external datasets or variability introduced by repeating model development.

Because we derived the patient-level target from available longitudinal EHR coding, incomplete, delayed, or temporally inconsistent subtype coding may have affected the assigned target. The ever-chronic rule assigned one target per patient and did not model migraine coding as a time-varying state. Patients with longer observation periods or more visits had more opportunities for a qualifying chronic migraine code and other clinical events to be recorded, so differences in coding opportunity may have influenced both the target value and the information available to the model. The code-history audit characterized patterns in the available records but did not estimate target misclassification against an external reference standard. In the exploratory visit-count analyses, the number of unique recorded dates was used as a proxy for documentation intensity. Because the visit-count strata differed in target-positive prevalence and patient composition, we treated their performance differences as descriptive and did not interpret them as causal effects of visit count, documentation intensity, or migraine severity.

Future work should include external validation in independent EHR cohorts and across institutions, as well as evaluation in other diseases and patient-level analysis tasks. Validation against chart-adjudicated diagnoses or alternative target-assignment rules is needed to assess the EHR-derived target. Future studies should also evaluate prospective designs that restrict model inputs to records available before a prespecified prediction time.

Taken together, these results support PaReGTA as a modular approach for generating fixed-dimensional, temporally aware patient-level representations for retrospective EHR analysis, while additional evaluation is needed across independent cohorts, diseases, and patient-level tasks.

\section*{Funding}

This work is supported by a grant from the Korea Institute for Advancement of Technology (KIAT), funded by the Ministry of Trade, Industry and Energy (MOTIE), Republic of Korea, through the Human Resource Development Program for Industrial Innovation (Global) (Grant No. RS-2024-00435815).

\section*{Role of the funder}

The funder had no role in the study design; data collection, analysis, or interpretation; preparation of the manuscript; or the decision to submit the manuscript for publication.

\section*{Disclosure of interest}

The authors report no conflict of interest.

\section*{Consent and Approval Statement}

This study used de-identified secondary data from the All of Us Research Program. All participants in the All of Us Research Program provided informed consent for the use of their data in research. This study was exempt from the requirement for institutional review board approval because it involved the secondary analysis of de-identified data.

\section*{Acknowledgements}

In the course of this study, we used ChatGPT 5.3, ChatGPT 5.4 Pro, and Claude Opus 4.6 for the following purposes: (i) coding assistance, (ii) language improvement, and (iii) literature classification. For coding assistance, we needed efficient code development to reduce the time required for experiments. Accordingly, we provided pseudocode and requested code implementation, or asked for reviews of code we had written. All code generated by LLMs was subjected to rigorous review by the authors. We also used LLMs for proofreading to improve readability and asked them to organize the references we had identified.

We gratefully acknowledge All of Us participants for their contributions, without whom this research would not have been possible. We also thank the National Institutes of Health’s All of Us Research Program for making available the participant data and migraine cohort examined in this study.

Dr. Kihyuk Yoon is a visiting postdoc at Georgia Tech. His visit is supported by a Korean Institute for Advancement of Technology (KIAT) grant funded by the Korean Government (MOTIE) (RS-2024-00435815, Human Resource Development Program for Industrial Innovation (Global)). The authors also thank Dr. Chiehyeon Lim for his valuable input on the manuscript.

\bibliographystyle{apacite}
\bibliography{draft_iise}

\appendix

\section{Data availability}

The EHR data used in this study contain protected health information, which should only be available under AoU policy. 

\section{Migraine case-study data preparation}
\label{app:migraine_reference_dates}

For each patient, auxiliary-date screening was initialized when at least one medication start record occurred within $\pm3$ days of a migraine-condition date. For each interval between successive migraine-condition dates, medication start dates occurring strictly more than three days after the earlier condition date and strictly more than three days before the later condition date were considered chronologically. Let $S_{\mathrm{current}}$ denote the current set of medication names and $S_d$ the set recorded at a candidate date $d$. The candidate date was retained when $|S_d|>1$, the absolute difference between the two set sizes was less than three, and fewer than three medications in $S_{\mathrm{current}}$ were absent from $S_d$. After a candidate date was retained, $S_d$ became the current medication set for the subsequent comparison.

We normalized condition, medication, and comorbidity intervals before constructing the patient-visit table. Records with both start and end dates missing were excluded; when only one endpoint was available, we used it for the missing endpoint. We removed exact duplicate records within each source domain before linkage. Migraine-condition dates and retained auxiliary medication dates formed the visit dates for the case study. A medication or comorbidity record was linked to every visit date that fell within its normalized interval, so the procedure did not use nearest-date exclusive assignment. Events assigned to the same constructed visit were organized within the corresponding visit-level row during preprocessing. After condition and auxiliary visit dates were combined, we did not additionally collapse distinct visit rows solely because they shared the same calendar date. Rows with neither medication nor comorbidity content were excluded, whereas rows with content in only one domain were retained for subsequent textualization.

\section{Operational ICD code sets}
\label{app_icd_codebook}

Tables~\ref{tab:migraine_icd_codes} and~\ref{tab:comorbidity_icd_codes} list the prespecified ICD-9-CM and ICD-10-CM codes used for the operational migraine and comorbidity categories, respectively.

\begingroup
\footnotesize
\setlength{\tabcolsep}{5pt}
\renewcommand{\arraystretch}{1.08}
\begin{longtable}{>{\raggedright\arraybackslash}p{0.56\linewidth}>{\raggedright\arraybackslash}p{0.18\linewidth}>{\raggedright\arraybackslash}p{0.16\linewidth}}
\caption{Prespecified ICD codes used for the operational migraine categories.}
\label{tab:migraine_icd_codes}\\
\toprule
\textbf{Category name} & \textbf{ICD} & \textbf{Code} \\
\midrule
\endfirsthead

\multicolumn{3}{l}{Table~\thetable\ continued from the previous page.}\\
\toprule
\textbf{Category name} & \textbf{ICD} & \textbf{Code} \\
\midrule
\endhead

\midrule
\multicolumn{3}{r}{Continued on the next page.}\\
\endfoot

\bottomrule
\endlastfoot

Migraine (all) & ICD-9-CM & 346 \\
Migraine (all) & ICD-10-CM & G43 \\
Chronic migraine without aura & ICD-9-CM & 346.7 \\
Chronic migraine without aura & ICD-9-CM & 346.70 \\
Chronic migraine without aura & ICD-9-CM & 346.71 \\
Chronic migraine without aura & ICD-9-CM & 346.72 \\
Chronic migraine without aura & ICD-9-CM & 346.73 \\
Chronic migraine without aura & ICD-10-CM & G43.7 \\
Chronic migraine without aura & ICD-10-CM & G43.70 \\
Chronic migraine without aura & ICD-10-CM & G43.701 \\
Chronic migraine without aura & ICD-10-CM & G43.709 \\
Chronic migraine without aura & ICD-10-CM & G43.71 \\
Chronic migraine without aura & ICD-10-CM & G43.711 \\
Chronic migraine without aura & ICD-10-CM & G43.719 \\
Menstrual migraine & ICD-9-CM & 346.4 \\
Menstrual migraine & ICD-9-CM & 346.40 \\
Menstrual migraine & ICD-9-CM & 346.41 \\
Menstrual migraine & ICD-9-CM & 346.42 \\
Menstrual migraine & ICD-9-CM & 346.43 \\
Menstrual migraine & ICD-10-CM & G43.82 \\
Menstrual migraine & ICD-10-CM & G43.821 \\
Menstrual migraine & ICD-10-CM & G43.829 \\
Menstrual migraine & ICD-10-CM & G43.83 \\
Menstrual migraine & ICD-10-CM & G43.831 \\
Menstrual migraine & ICD-10-CM & G43.839 \\
Migraine with aura & ICD-9-CM & 346.0 \\
Migraine with aura & ICD-9-CM & 346.00 \\
Migraine with aura & ICD-9-CM & 346.01 \\
Migraine with aura & ICD-9-CM & 346.02 \\
Migraine with aura & ICD-9-CM & 346.03 \\
Migraine with aura & ICD-10-CM & G43.1 \\
Migraine with aura & ICD-10-CM & G43.10 \\
Migraine with aura & ICD-10-CM & G43.101 \\
Migraine with aura & ICD-10-CM & G43.109 \\
Migraine with aura & ICD-10-CM & G43.11 \\
Migraine with aura & ICD-10-CM & G43.111 \\
Migraine with aura & ICD-10-CM & G43.119 \\
Migraine without aura & ICD-9-CM & 346.1 \\
Migraine without aura & ICD-9-CM & 346.10 \\
Migraine without aura & ICD-9-CM & 346.11 \\
Migraine without aura & ICD-9-CM & 346.12 \\
Migraine without aura & ICD-9-CM & 346.13 \\
Migraine without aura & ICD-10-CM & G43.0 \\
Migraine without aura & ICD-10-CM & G43.00 \\
Migraine without aura & ICD-10-CM & G43.001 \\
Migraine without aura & ICD-10-CM & G43.009 \\
Migraine without aura & ICD-10-CM & G43.01 \\
Migraine without aura & ICD-10-CM & G43.011 \\
Migraine without aura & ICD-10-CM & G43.019 \\
Migraine, unspecified & ICD-9-CM & 346.9 \\
Migraine, unspecified & ICD-9-CM & 346.90 \\
Migraine, unspecified & ICD-9-CM & 346.91 \\
Migraine, unspecified & ICD-9-CM & 346.92 \\
Migraine, unspecified & ICD-9-CM & 346.93 \\
Migraine, unspecified & ICD-10-CM & G43.9 \\
Migraine, unspecified & ICD-10-CM & G43.90 \\
Migraine, unspecified & ICD-10-CM & G43.901 \\
Migraine, unspecified & ICD-10-CM & G43.909 \\
Migraine, unspecified & ICD-10-CM & G43.91 \\
Migraine, unspecified & ICD-10-CM & G43.911 \\
Migraine, unspecified & ICD-10-CM & G43.919 \\
Other migraine & ICD-9-CM & 346.8 \\
Other migraine & ICD-9-CM & 346.80 \\
Other migraine & ICD-10-CM & G43.8 \\
Other migraine & ICD-10-CM & G43.80 \\
Other migraine & ICD-10-CM & G43.801 \\
Other migraine & ICD-10-CM & G43.809 \\
Other migraine & ICD-10-CM & G43.81 \\
Other migraine & ICD-10-CM & G43.811 \\
Other migraine & ICD-10-CM & G43.819 \\
\end{longtable}
\endgroup

\begingroup
\footnotesize
\setlength{\tabcolsep}{5pt}
\renewcommand{\arraystretch}{1.08}
\begin{longtable}{>{\raggedright\arraybackslash}p{0.56\linewidth}>{\raggedright\arraybackslash}p{0.18\linewidth}>{\raggedright\arraybackslash}p{0.16\linewidth}}
\caption{Prespecified ICD codes used for the operational comorbidity categories.}
\label{tab:comorbidity_icd_codes}\\
\toprule
\textbf{Category name} & \textbf{ICD} & \textbf{Code} \\
\midrule
\endfirsthead

\multicolumn{3}{l}{Table~\thetable\ continued from the previous page.}\\
\toprule
\textbf{Category name} & \textbf{ICD} & \textbf{Code} \\
\midrule
\endhead

\midrule
\multicolumn{3}{r}{Continued on the next page.}\\
\endfoot

\bottomrule
\endlastfoot

Depression & ICD-10-CM & F32.A \\
Depression & ICD-10-CM & Z13.31 \\
Depression & ICD-10-CM & Z13.32 \\
Depression & ICD-10-CM & F53.0 \\
Depression & ICD-9-CM & 290.13 \\
Depression & ICD-9-CM & V79.0 \\
Vulvodynia & ICD-9-CM & 625.7 \\
Vulvodynia & ICD-10-CM & N94.81 \\
Myalgic encephalomyelitis / chronic fatigue syndrome & ICD-10-CM & G93.32 \\
Endometriosis & ICD-9-CM & 617 \\
Endometriosis & ICD-10-CM & N80 \\
Anxiety & ICD-9-CM & 300.0 \\
Chronic tension-type headache & ICD-9-CM & 339.12 \\
Chronic tension-type headache & ICD-10-CM & G44.22 \\
Temporomandibular joint disorders & ICD-9-CM & 524.6 \\
Temporomandibular joint disorders & ICD-10-CM & M26.6 \\
Interstitial cystitis & ICD-9-CM & 595.1 \\
Interstitial cystitis & ICD-10-CM & N30.1 \\
Diabetes mellitus & ICD-9-CM & 250 \\
Diabetes mellitus & ICD-10-CM & E10 \\
Diabetes mellitus & ICD-10-CM & E11 \\
Diabetes mellitus & ICD-10-CM & Z83.3 \\
Alcohol-related conditions & ICD-9-CM & 305.0 \\
Alcohol-related conditions & ICD-10-CM & F10 \\
Alcohol-related conditions & ICD-10-CM & F10.1 \\
Alcohol-related conditions & ICD-10-CM & F10.2 \\
Alcohol-related conditions & ICD-10-CM & K70 \\
Post-traumatic stress disorder (PTSD) & ICD-10-CM & F43.1 \\
Dementia & ICD-9-CM & 294.2 \\
Dementia & ICD-10-CM & F03 \\
Irritable bowel syndrome & ICD-9-CM & 564.1 \\
Irritable bowel syndrome & ICD-10-CM & K58 \\
Fibromyalgia & ICD-10-CM & M79.7 \\
Concussion & ICD-9-CM & 850 \\
Concussion & ICD-10-CM & S06.0 \\
Hypertension & ICD-9-CM & 401 \\
Hypertension & ICD-10-CM & I10 \\
Insomnia & ICD-10-CM & G47.0 \\
Insomnia & ICD-10-CM & F51.0 \\
Insomnia & ICD-9-CM & 780.51 \\
Insomnia & ICD-9-CM & 780.52 \\
\end{longtable}
\endgroup

\section{Environment} 
\label{app_environment}

Due to data sensitivity, AoU data cannot be exported to personal environments, so all implementation and analyses have to be performed within the AoU workspace. Our workspace was provisioned with 4 CPUs and a single NVIDIA Tesla T4 GPU. We implemented the framework in Python 3.10.16. Key package versions were as follows: \texttt{numpy} 1.23.5, \texttt{pycaret} 3.1.0, \texttt{pytorch} 2.6.0, and \texttt{transformers} 4.51.3. We used \texttt{pycaret} to train a broad set of machine-learning models and to apply standard preprocessing, \texttt{pytorch} as the deep-learning framework, and \texttt{transformers} to access pretrained models from Hugging Face. 

\section{Experimental setting}
\label{app_ex_setting}

This appendix details the hyperparameter configurations used for target-free SimCSE adaptation of the sentence-embedding model and label-supervised training of the attention pooler. For all experiments except T-LSTM and RETAIN, 42 was used as the random seed.

\begin{table}[t]
\tbl{Hyperparameters for SimCSE fine-tuning.}
{\begin{tabular}{lll} 
\toprule
Hyperparameter & Value & Description \\
\midrule
Epochs & 1 & Number of training epochs \\
Batch size & 128 & Number of sentence pairs per batch \\
Learning rate & $2 \times 10^{-5}$ & AdamW optimizer learning rate \\
Max training samples & 200,000 & Maximum unique sentences for training \\
Warmup ratio & 0.05 & Proportion of steps for LR warmup \\
Gradient clipping & 1.0 & Maximum gradient norm \\
Gradient accumulation & 4 & Steps to accumulate gradients \\
Mixed precision (FP16) & True & Use half-precision training \\
Max sequence length & 256 & Maximum tokens per input \\
\bottomrule
\end{tabular}}
\label{tab:simcse_params}
\end{table}

\subsection{SimCSE fine-tuning}

We fine-tuned GTE-base-v1.5 (\texttt{Alibaba-NLP/gte-base-en-v1.5}) using the identical-text \texttt{MultipleNegativesRankingLoss} objective described in Methods \citep{reimers_sbert_2019,gao_simcse_2021}. Table~\ref{tab:simcse_params} summarizes the hyperparameters used for SimCSE fine-tuning.

\begin{table}[t]
\tbl{Hyperparameters for hybrid temporal pooling.}
{\begin{tabular}{lll} 
\toprule
Hyperparameter & Value & Description \\
\midrule
Pooling method & hybrid & Time-weighted + attention pooling \\
$\gamma$ & 0.05 & Time decay factor \\
$\delta$ mode & log1p & Time gap transformation \\
$\alpha$ & 0.5 & Attention vs.\ time-weighted weight \\
\bottomrule
\end{tabular}}
\label{tab:pooling_params}
\end{table}

\subsection{Hybrid temporal pooling}

Table~\ref{tab:pooling_params} presents the pooling configuration for aggregating visit-level embeddings.

\subsection{Label-supervised attention-pooler training}

\begin{table}[t]
\tbl{Hyperparameters for label-supervised attention-pooler training.}
{\begin{tabular}{lll} 
\toprule
Hyperparameter & Value & Description \\
\midrule
Epochs & 3 & Number of training epochs \\
Batch size (triplets) & 32 & Triplets per batch \\
Learning rate & $1 \times 10^{-3}$ & AdamW optimizer learning rate \\
Weight decay & $1 \times 10^{-4}$ & L2 regularization coefficient \\
Triplet margin & 0.2 & Margin for triplet loss \\
Hidden dimension & 96 & Hidden layer size in attention MLP \\
$\delta$ mode & log1p & Time gap transformation \\
Maximum triplets per reference patient & 4 & Maximum positive patients per reference patient \\
Output normalization & True & L2-normalize pooled vectors for triplet loss \\
Random seed & 42 & Seed for reproducibility \\
\bottomrule
\end{tabular}}
\label{tab:attn_params}
\end{table}

Table~\ref{tab:attn_params} summarizes the hyperparameters used for label-supervised attention-pooler training. The implementation used only patients present in both the designated training visit-embedding data and the corresponding training-label file. The poolers were trained for three fixed epochs without validation-based early stopping or best-checkpoint selection, and the final parameter states were reused without further fitting for held-out representation generation. Separate pooler instances and checkpoints were used for the medication and comorbidity modalities.

\subsection{Details on baseline models}
\label{subsec:details_baseline}

One-hot encoding provided a sparse patient-level indicator representation, whereas the Bag-of-Codes (BoC) baseline used patient-level event-occurrence counts. For MedBERT and CEHR-BERT, we kept the patient-level target separate from the input and constructed medication and comorbidity event-token vocabularies using only the inner training subset, retaining tokens observed at least five times. We omitted visits with no retained event tokens. Within each retained visit, we deduplicated and alphabetically sorted medication and comorbidity tokens within their respective domains and placed medication tokens before comorbidity tokens. We limited each patient sequence to 512 tokens; when a sequence exceeded this limit, we removed the oldest event tokens while preserving \texttt{[CLS]} and retained the most recent history.

MedBERT represented each patient using \texttt{[CLS]}, visit-level event tokens separated by \texttt{[SEP]}, and visit-segment identifiers, without explicit age, elapsed-time, or inter-visit gap features. CEHR-BERT additionally used \texttt{[VS]} and \texttt{[VE]} visit-boundary tokens, alternating visit-segment identifiers, age-at-visit and elapsed-time features, and gap-derived \texttt{ATT} tokens. We trained both encoders from random initialization within the study cohort rather than importing external pretrained checkpoints. Using only the inner training subset, we performed masked language modeling for MedBERT and masked language modeling together with \texttt{ATT}-bucket prediction for CEHR-BERT, and then fine-tuned each encoder for patient-level binary classification. We selected the checkpoint with the highest validation AUC and evaluated each selected model once on the fixed test cohort.

T-LSTM and RETAIN used the same chronologically ordered sequence of 46 aligned variables: 15 personal and demographic, 14 medication, and 17 comorbidity variables. We retained same-day rows in their stable source order rather than merging them. When a sequence exceeded 512 rows, we removed the oldest rows and retained the most recent 512. We then recomputed the elapsed days between consecutive retained rows, set the first elapsed-time value to 0, and constrained subsequent values to 0--3,650 days. T-LSTM used these elapsed-time values in its memory-decay mechanism, whereas RETAIN used the 46 visit-row features without an explicit elapsed-time input. Because visit-sequence lengths varied substantially across patients, we used length-bucket batching for both models to group patients with similar sequence lengths within the same batch and address the resulting training instability. We selected the model configurations using validation results only, trained the selected configurations with seeds 0--4, and evaluated each run on the same fixed held-out test cohort. We summarized fixed-test performance as the mean and sample standard deviation across the five seeds.

\subsection{Details on hyperparameters of PaReGTA}
\label{subsec:detail_hyperparameter}

For the PaReGTA-Gap reference configuration, we prespecified $\gamma=0.05$ as a small positive decay coefficient applied to log-transformed elapsed days, with the intention of emphasizing recent visits while retaining contributions from older records. We prespecified $\alpha=0.50$ as a neutral starting value that assigned equal contributions to the attention-pooled and time-weighted representations. Here, $\alpha$ denotes the attention contribution, whereas $1-\alpha$ denotes the time-weighted contribution. These values were fixed design constants and were not selected through separate fine-tuning, validation-based optimization, or test-set comparison. Appendix Table~\ref{tab:pooling_sensitivity} reports a non-exhaustive sensitivity analysis of completed alternative settings.

\section{Additional experimental results}

\subsection{Intentional label-inclusion positive control}
\label{subsec:label_positive_control}

Table~\ref{table_label_positive_control} presents point estimates from the intentional label-inclusion positive control. On the same fixed held-out test cohort, the label-included configuration achieved an AUC of 0.9741, an accuracy of 0.9850, and an F1 score of 0.9609, compared with 0.9501, 0.9229, and 0.8147, respectively, for the Label-excluded PaReGTA-Gap reference input.

\begin{table}
\tbl{Descriptive label-inclusion positive control using PaReGTA-Gap with LightGBM on the fixed held-out test set ($n=7{,}818$).}
{\begin{tabular}{lccc}
\toprule
Configuration &
AUC &
Accuracy &
F1 \\
\midrule
Primary PaReGTA-Gap input &
0.9501 &
0.9229 &
0.8147 \\
Label-included positive control &
0.9741 &
0.9850 &
0.9609 \\
\bottomrule
\end{tabular}}
\label{table_label_positive_control}
\end{table}

The higher point estimates are consistent with the model using the deliberately supplied label-bearing field. This positive control was not a candidate model and does not establish that the primary input excluded the label; that exclusion was established by the input contract and fail-fast checks described in Methods. The analysis also does not rule out indirect label-related signals in the remaining EHR data.

\subsection{Fixed-window baseline}

\begin{table}
\tbl{Performance of window encoding and LightGBM.}
{
\begin{tabular}{lccc}
\toprule
Model &
AUC &
accuracy &
F1 \\
\midrule
3-month window + LightGBM &
0.8495 &
0.8509 &
0.5460 \\
6-month window + LightGBM &
0.8485 &
0.8451 &
0.5230 \\
\bottomrule
\end{tabular}
}
\label{table_window}
\end{table}

For each fixed-window baseline, we measured time backward from each patient's last retained record date, assigned each medication and comorbidity occurrence to a non-overlapping 3- or 6-month interval, and counted occurrences within each interval. The look-back horizon was determined from the 95th percentile of follow-up in the outer training pool, rounded upward to a complete interval, and constrained to 24--48 months; records older than this horizon were retained in a single plus-bin. The medication and comorbidity type lists used to define the feature columns were enumerated from the full preprocessed analytic table before the PID partitions were applied. All fitted preprocessing and classifier parameters were estimated from the outer training pool.

\subsection{Sensitivity of PaReGTA to hyperparameters}
\label{subsec:hyperparameter_experiment}

We conducted a non-exhaustive sensitivity analysis using the fixed PaReGTA-Gap--LightGBM pipeline on the same test cohort of 7,818 patients. The analysis included the prespecified primary configuration and 10 completed alternative settings. These fixed-test results were not used to select or replace the primary configuration and are therefore interpreted as a sensitivity analysis rather than hyperparameter optimization.

\begin{table}
\tbl{Sensitivity of PaReGTA-Gap--LightGBM to the temporal-decay parameter $\gamma$ and attention contribution $\alpha$ on the fixed test cohort.}
{
\begin{tabular}{ccccc}
\toprule
$\gamma$ &
$\alpha$ &
AUC &
accuracy &
F1 \\
\midrule
0.200 & 0.50 & 0.9435 & 0.9027 & 0.7751 \\
0.100 & 0.50 & 0.9507 & 0.9221 & 0.8123 \\
0.100 & 0.25 & 0.9496 & 0.9179 & 0.8041 \\
0.050 & 0.90 & 0.9482 & 0.9190 & 0.8073 \\
0.050 & 0.75 & 0.9512 & 0.9253 & 0.8215 \\
0.050 & 0.50 & 0.9501 & 0.9229 & 0.8147 \\
0.050 & 0.25 & 0.9497 & 0.9212 & 0.8108 \\
0.050 & 0.10 & 0.9479 & 0.9204 & 0.8055 \\
0.010 & 0.75 & 0.9523 & 0.9252 & 0.8213 \\
0.010 & 0.50 & 0.9531 & 0.9248 & 0.8185 \\
0.005 & 0.50 & 0.9533 & 0.9236 & 0.8148 \\
\bottomrule
\end{tabular}
}
\label{tab:pooling_sensitivity}
\end{table}

Across the completed settings, AUC ranged from 0.9435 to 0.9533, accuracy from 0.9027 to 0.9253, and F1 from 0.7751 to 0.8215. Results were similar across most examined settings, although the largest tested decay coefficient on the log-transformed elapsed-day scale, $\gamma=0.20$, produced lower accuracy and F1. The highest observed AUC occurred at $\gamma=0.005$ and $\alpha=0.50$, whereas the highest accuracy and F1 occurred at $\gamma=0.05$ and $\alpha=0.75$; in the latter setting, $\alpha=0.75$ assigned greater weight to the attention-pooled representation than to the time-weighted representation. These descriptive maxima did not replace the prespecified primary configuration.

\subsection{Exploratory analyses by visit count}
\label{subsec:visit_count_analysis}

We defined visit count as the number of distinct normalized recorded dates per patient. We applied a target-blind three-quantile partition to the fixed-test visit-count distribution using visit counts alone. Because tied integer counts produced duplicate quantile boundaries, the partition yielded two strata: 1--7 visits and at least 8 visits. The 8-visit boundary was therefore an exploratory operational boundary rather than a clinical cutoff, prespecified threshold, or training-derived threshold, and it was not selected using the target or model performance.

The first design was a post-hoc subgroup evaluation. We retained the existing fixed-test predictions from the PaReGTA-Gap + LightGBM model developed without visit-count stratification and recalculated AUC, accuracy, and F1 within each stratum. No subgroup-specific refitting was performed. Corresponding post-hoc subgroup predictions were not available for T-LSTM or RETAIN, so these models were not added to this analysis.

\begin{table}
\tbl{Post-hoc performance of the PaReGTA-Gap + LightGBM model developed without visit-count stratification, evaluated using its existing fixed-test predictions.}
{\begin{tabular}{lccc}
\toprule
Evaluation group &
AUC &
Accuracy &
F1 \\
\midrule
All fixed test &
0.9501 &
0.9229 &
0.8147 \\
1--7 visits &
0.9268 &
0.9405 &
0.7459 \\
At least 8 visits &
0.9539 &
0.8869 &
0.8564 \\
\bottomrule
\end{tabular}}
\label{table_visit_count_posthoc}
\end{table}

Table~\ref{table_visit_count_posthoc} shows that the stratum with at least 8 visits had higher AUC and F1 but lower accuracy than the 1--7-visit stratum under the fixed-prediction evaluation.

The second design was cohort-specific retraining. We applied the same 1--7-visit and at-least-8-visit rule to the frozen outer patient assignments. Within each stratum, we divided the stratum-specific outer training pool into train-fit and validation subsets using a target-stratified 90/10 split with seed 0. PaReGTA-Gap preprocessing and downstream classifier fitting used the combined train-fit and validation patients, whereas T-LSTM and RETAIN used the train-fit subset for fitting and the validation subset for model selection. We then evaluated each newly fitted model on the corresponding fixed-test subgroup without creating a new outer partition. T-LSTM and RETAIN were trained with seeds 0--4, and their results were summarized as the mean $\pm$ sample standard deviation across the five seeds.

\begin{table}
\tbl{Performance after cohort-specific retraining by visit-count stratum. The all-visit rows reproduce the full-cohort reference results and were not obtained by stratum-specific retraining. PaReGTA-Gap + LightGBM values are fixed-test point estimates, whereas T-LSTM and RETAIN values are means $\pm$ sample standard deviations across seeds 0--4.}
{\begin{tabular}{llccc}
\toprule
Visit cohort &
Model &
AUC &
Accuracy &
F1 \\
\midrule
\multirow{3}{*}{All visits (reference)} &
PaReGTA-Gap + LightGBM &
0.9501 &
0.9229 &
0.8147 \\
&
T-LSTM &
$0.9447 \pm 0.0011$ &
$0.9093 \pm 0.0042$ &
$0.7975 \pm 0.0053$ \\
&
RETAIN &
$0.8559 \pm 0.0024$ &
$0.7632 \pm 0.0114$ &
$0.5544 \pm 0.0078$ \\
\midrule
\multirow{3}{*}{1--7 visits} &
PaReGTA-Gap + LightGBM &
0.9221 &
0.9434 &
0.7595 \\
&
T-LSTM &
$0.9218 \pm 0.0024$ &
$0.9223 \pm 0.0048$ &
$0.7136 \pm 0.0120$ \\
&
RETAIN &
$0.8465 \pm 0.0031$ &
$0.7828 \pm 0.0338$ &
$0.4352 \pm 0.0077$ \\
\midrule
\multirow{3}{*}{At least 8 visits} &
PaReGTA-Gap + LightGBM &
0.9530 &
0.8862 &
0.8522 \\
&
T-LSTM &
$0.9356 \pm 0.0050$ &
$0.8743 \pm 0.0075$ &
$0.8354 \pm 0.0086$ \\
&
RETAIN &
$0.8017 \pm 0.0052$ &
$0.7233 \pm 0.0344$ &
$0.6530 \pm 0.0065$ \\
\bottomrule
\end{tabular}}
\label{table_visit_count_retraining}
\end{table}

The fixed test cohort of 7,818 patients had a median of 3 distinct recorded visit dates and a maximum of 1,123. The 1--7-visit stratum comprised 5,244 patients, with a median of 1 visit; 601 patients were target-positive and 4,643 were target-negative, corresponding to a target-positive prevalence of 11.5\%. The stratum with at least 8 visits comprised 2,574 patients, with a median of 20 visits; 948 patients were target-positive and 1,626 were target-negative, corresponding to a target-positive prevalence of 36.8\%.

Table~\ref{table_visit_count_retraining} reports the cohort-specific retraining results. In the 1--7-visit stratum, the AUC values were 0.9221 for PaReGTA-Gap + LightGBM, $0.9218 \pm 0.0024$ for T-LSTM, and $0.8465 \pm 0.0031$ for RETAIN. In the stratum with at least 8 visits, the corresponding values were 0.9530, $0.9356 \pm 0.0050$, and $0.8017 \pm 0.0052$, respectively. PaReGTA-Gap + LightGBM and T-LSTM both had higher AUC and F1 but lower accuracy in the stratum with at least 8 visits than in the 1--7-visit stratum. These are descriptive numerical comparisons; the all-visit rows serve only as references and do not represent additional stratum-specific fits.

\subsection{Fixed-test bootstrap analysis}
\label{subsec:fixed_test_bootstrap}

We quantified patient-sampling uncertainty for models with available patient-level fixed-test predictions using a nonparametric bootstrap. For each model, we generated 1,000 bootstrap samples by drawing 7,818 patients with replacement from the fixed held-out test cohort and recalculated AUC, accuracy, and F1 from the corresponding fixed predictions. We did not retrain any model, refit preprocessing transformations, repeat the data partition, or alter the training, validation, and test assignments. We defined each 95\% percentile confidence interval using the 2.5th and 97.5th percentiles of the 1,000 bootstrap estimates; the reported point estimate was calculated from the original fixed-test predictions.

The fixed-prediction analysis included PaReGTA-Gap + LightGBM, one-hot encoding + LightGBM, BoC + LightGBM, MedBERT, CEHR-BERT, T-LSTM, and RETAIN. For T-LSTM and RETAIN, which were trained with seeds 0--4, we used the fixed-test predictions from the seed with the highest accuracy on the held-out test set.

For the paired comparison, we aligned predictions by patient identifier and target value and applied the same resampled patient indices to PaReGTA-Gap + LightGBM and T-LSTM. We calculated each difference as PaReGTA-Gap + LightGBM minus T-LSTM and obtained the paired 95\% confidence interval from the 2.5th and 97.5th percentiles of the 1,000 paired differences. T-LSTM was selected because it achieved the best performance among the evaluated baseline models.

\begin{table}
\tbl{Fixed-test point estimates and 95\% percentile bootstrap confidence intervals based on 1,000 patient-level resamples of fixed predictions.}
{\begin{tabular}{lccc}
\toprule
Model &
AUC &
Accuracy &
F1 \\
\midrule
PaReGTA-Gap + LightGBM &
0.9501 (0.9436--0.9568) &
0.9229 (0.9170--0.9286) &
0.8147 (0.8002--0.8287) \\
One-hot encoding + LightGBM &
0.7634 (0.7491--0.7775) &
0.8402 (0.8322--0.8486) &
0.4581 (0.4346--0.4849) \\
BoC + LightGBM &
0.8354 (0.8243--0.8463) &
0.8473 (0.8396--0.8547) &
0.5162 (0.4940--0.5397) \\
MedBERT &
0.8600 (0.8502--0.8694) &
0.7970 (0.7874--0.8058) &
0.5831 (0.5642--0.6018) \\
CEHR-BERT &
0.9173 (0.9091--0.9241) &
0.8175 (0.8084--0.8259) &
0.6566 (0.6398--0.6724) \\
T-LSTM &
0.9428 (0.9361--0.9501) &
0.9121 (0.9060--0.9184) &
0.8025 (0.7882--0.8162) \\
RETAIN &
0.8573 (0.8468--0.8671) &
0.7739 (0.7639--0.7829) &
0.5602 (0.5412--0.5783) \\
\bottomrule
\end{tabular}}
\label{table_fixed_test_bootstrap_ci}
\end{table}

Table~\ref{table_fixed_test_bootstrap_ci} presents the fixed-test point estimates and corresponding 95\% percentile bootstrap confidence intervals. These intervals quantify patient-sampling variability conditional on the fixed prediction sets and do not incorporate uncertainty from model retraining, data repartitioning, or post-hoc representation and classifier selection.

\begin{table}
\tbl{Paired fixed-test differences between PaReGTA-Gap + LightGBM and T-LSTM. Differences are defined as PaReGTA-Gap + LightGBM minus T-LSTM; the 95\% confidence intervals are percentile intervals from 1,000 paired patient-level bootstrap resamples.}
{\begin{tabular}{lcc}
\toprule
Metric &
Paired difference &
95\% confidence interval \\
\midrule
AUC &
+0.0072 &
+0.0024--+0.0124 \\
Accuracy &
+0.0107 &
+0.0056--+0.0157 \\
F1 &
+0.0122 &
+0.0007--+0.0231 \\
\bottomrule
\end{tabular}}
\label{table_paired_bootstrap_tlstm}
\end{table}

The paired differences were positive for AUC, accuracy, and F1, and all three percentile confidence intervals excluded zero. These results indicate modest but consistently positive differences in favor of PaReGTA-Gap + LightGBM under the evaluated fixed-test resampling procedure and should not be interpreted as evidence of clinical significance or general superiority across datasets and tasks.

\end{document}